\documentclass[journal]{IEEEtran}

\makeatletter
\def\endthebibliography{%
  \def\@noitemerr{\@latex@warning{Empty `thebibliography' environment}}%
  \endlist
}
\makeatother

\usepackage[applemac]{inputenc}

\usepackage{graphicx}
\usepackage{color}

\usepackage{cite}
\usepackage{hyperref}
\usepackage{url}

\usepackage[cmex10]{amsmath}
\usepackage{empheq}
\usepackage{amssymb,stmaryrd}
\usepackage{bm}
\usepackage{multirow}
\usepackage{makecell}

\usepackage{stfloats}
\usepackage{array}
\usepackage{wasysym}

\usepackage{footmisc}

\usepackage[ruled,vlined]{algorithm2e}

\usepackage{booktabs}
\usepackage{subfigure}

\begin{document}

\title{3D Consistent \& Robust Segmentation of Cardiac Images by Deep Learning with Spatial Propagation}

\author{Qiao Zheng, Herv\'{e} Delingette, Nicolas Duchateau, Nicholas Ayache
\thanks{Qiao Zheng, Herv\'{e} Delingette and Nicholas Ayache are with Universit\'{e} C\^{o}te d'Azur, Inria, France.}
\thanks{Nicolas Duchateau is with CREATIS, CNRS UMR 5220, INSERM U1206, France.}
\thanks{Address for correspondence: Qiao Zheng, Inria Asclepios, 2004 route des Lucioles BP 93
06902 Sophia Antipolis Cedex, France. Tel: +33.49238.5024; Fax +33.49238.7669.}
\thanks{Email: qiao.zheng@inria.fr}
\thanks{~}
}


\maketitle

\begin{abstract}
We propose a method based on deep learning to perform cardiac segmentation on short axis MRI image stacks iteratively from the top slice (around the base) to the bottom slice (around the apex). At each iteration, a novel variant of U-net is applied to propagate the segmentation of a slice to the adjacent slice below it. In other words, the prediction of a segmentation of a slice is dependent upon the already existing segmentation of an adjacent slice. 3D-consistency is hence explicitly enforced. The method is trained on a large database of 3078 cases from UK Biobank. It is then tested on 756 different cases from UK Biobank and three other state-of-the-art cohorts (ACDC with 100 cases, Sunnybrook with 30 cases, RVSC with 16 cases). Results comparable or even better than the state-of-the-art in terms of distance measures are achieved. They also emphasize the assets of our method, namely enhanced spatial consistency (currently neither considered nor achieved by the state-of-the-art), and the generalization ability to unseen cases even from other databases.

\end{abstract}


\IEEEpeerreviewmaketitle

\section{Introduction}

The manual segmentation of cardiac images is tedious and time-consuming, which is even more critical given the new availability of huge databases (e.g. UK Biobank \cite{Petersen:2016}). Magnetic resonance imaging (MRI) is widely used by cardiologists. Yet MRI is challenging to segment due to its anisotropic resolution with somewhat distant 2D slices which might be misaligned. There is hence a great need for automated and accurate cardiac MRI segmentation methods.

In recent years, many state-of-the-art cardiac segmentation methods are based on deep learning and substantially overcome the performance of previous methods. Currently, they dominate various cardiac segmentation challenges. For instance, in the Automatic Cardiac Diagnosis Challenge \footnote{\url{https://www.creatis.insa-lyon.fr/Challenge/acdc/}. Accessed September 15 2017} (ACDC) of MICCAI 2017, 9 out of the 10 cardiac segmentation methods were based on deep learning. In particular, the 8 best-ranked methods were all deep learning ones. Deep learning methods can be roughly divided into to 2 classes: 2D methods, which segment each slice independently (i.e.\cite{Winther:2017}, \cite{Tran:2016}, \cite{Baumgartner:2017}), and 3D methods, which segment multiple slices together as a volume (i.e.\cite{Isensee:2017}, \cite{Baumgartner:2017}). 2D methods are popular because they are lightweight and require less data for training. But as no 3D context is taken into consideration, they might hardly maintain the 3D-consistency between the segmentation of different slices, and even fail on ``difficult" slices. For example, the 2D method used in \cite{Tran:2016} achieves state-of-the-art segmentation on several widely used datasets but makes the most prominent errors in apical slices and even fails to detect the presence of the heart. 

On the other hand, 3D methods should theoretically be robust to these issues. But in \cite{Baumgartner:2017}, with experiments on the ACDC dataset, the authors found that all the 2D approaches they proposed consistently outperformed the 3D method being considered. In fact, 3D methods have some significant shortcomings. First, using 3D data drastically reduces the number of training images. Second, 3D methods mostly rely on 3D convolution. Yet border effects from 3D convolution may compromise the information in intermediate representations of the neural networks. Third, 3D methods require far more GPU memory. Therefore, substantial downsampling of data is often necessary for training and prediction, which causes loss of information.

One possible way to combine the strengths of 2D and 3D methods is to use recurrent neural networks. In \cite{Poudel:2016}, the authors merge U-Net \cite{Ronneberger:2015} and a recurrent unit into a neural network to process all slices in the same stack, arranging the slices from the base to the apex. Information from the slices already segmented in the stack is preserved in the recurrent unit and used as context while segmenting the current slice. Comparisons in \cite{Poudel:2016} prove that this contextual information is helpful to achieve better segmentation. However, the approaches based on recurrent neural networks are still limited. On the one hand, as the slice thickness (usually 5 to 10mm) is often very large compared to the slice resolution (usually 1 to 2mm), the correlation between slices is low except for adjacent slices. Thus, considering all slices at once may not be optimal. On the other hand, the prediction on each slice made by a recurrent neural network does not depend on an existing prediction. With this setting, the automatic segmentation is remarkably different from the procedure of human experts. As presented in \cite{Suinesiaputra:2015}, human experts are very consistent in the sense that the intra-observer variability is low; yet the inter-observer variability is high, as segmentation bias varies remarkably between human experts. Hence in general, for given a slice, there is no a unique correct segmentation. But human operators still maintain consistency in their predictions respectively. Being inspired by these facts, we adopt a novel perspective: we train our networks to explicitly maintain the consistency between the current segmentation and the already predicted segmentation on an adjacent slice. We do not assume that there is a unique correct segmentation. Instead, the prediction for the current slice explicitly depends on another previously predicted segmentation.

Another possible method to improve segmentation consistency is to incorporate anatomical prior knowledge into neural networks. In \cite{Oktay:2018}, the segmentation models are trained to follow the cardiac anatomical properties via a learned representation of the 3D shape. While adopting novel training procedure, this method is based on 3D convolution neural networks for segmentation. So the issues of 3D methods discussed above still exist.

In this paper, we propose a novel method based on deep learning to perform cardiac segmentation. Our main contribution is threefold:\\
\textbullet \ The spatial consistency in cardiac segmentation is barely addressed in general, while this is a remarkable aspect of human expertise. Our method explicitly provides spatially consistent results by propagating the segmentations across slices. This is a novel perspective, as we do not assume the existence of a unique correct segmentation, and the prediction of the current segmentation depends on the segmentation of the previous slice. \\
\textbullet \ After training our method with a large dataset, we demonstrate its robustness and generalization ability on a large number of unseen cases from the same cohort as well as from other reference databases. These aspects are crucial for the application of a segmentation model in general, yet have not yet been explored before. \\
\textbullet \ Most segmentation methods proceed in a 2D manner to benefit from more training samples and higher training speed in comparison with 3D methods. In contrast, we proposed an original approach that keeps the computational assets of 2D methods but still addresses key 3D issues. \\
We hence believe in its potential impact on the community \footnote{The code and the models are available in this repository: https://github.com/julien-zheng/CardiacSegmentationPropagation}.


\begin{figure*}
\centering
\subfigure{
\includegraphics[width=4.2cm, height=5.5cm]{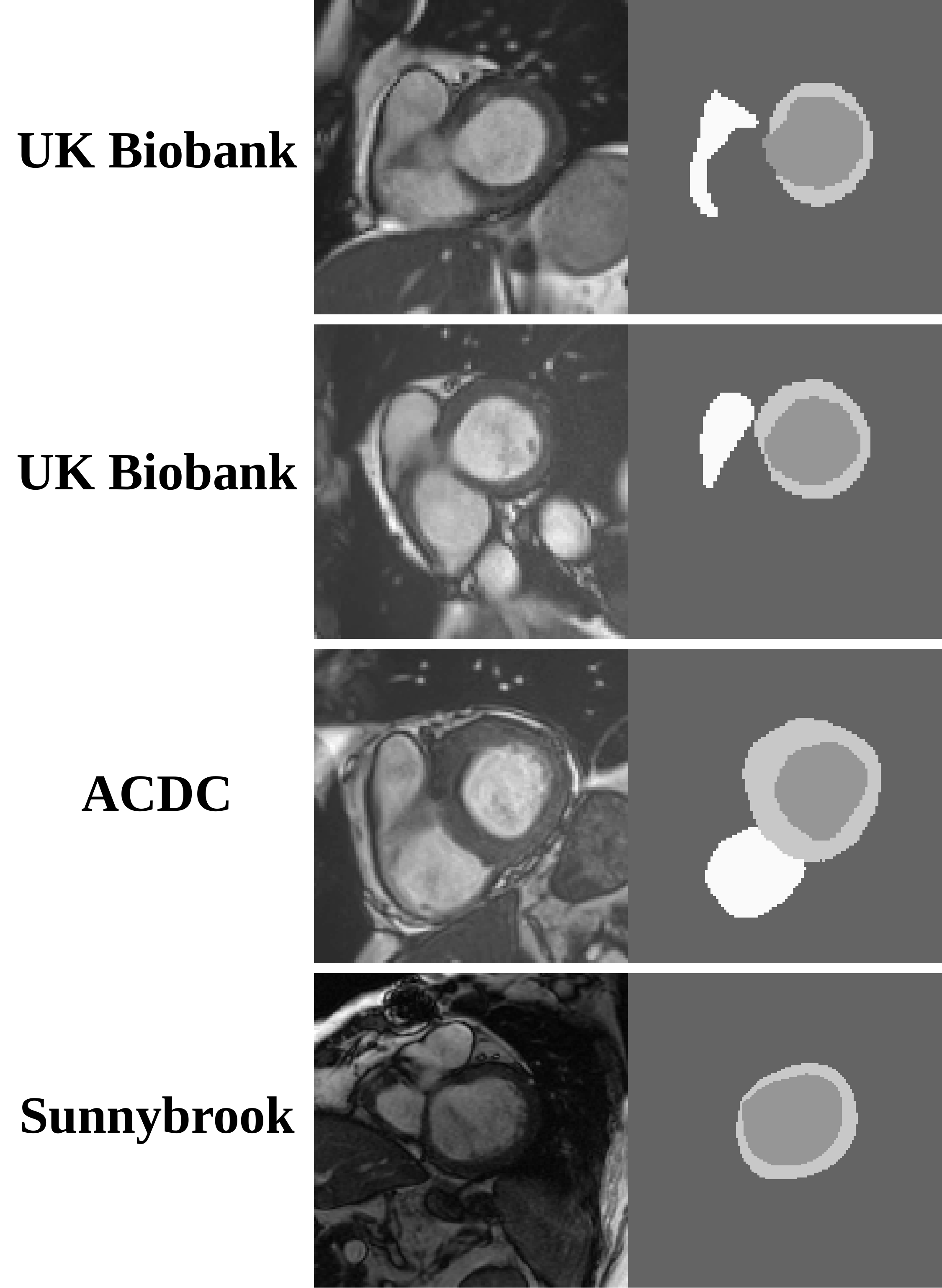}}\quad\quad\quad\quad
\subfigure{
\includegraphics[width=10.5cm, height=5.5cm]{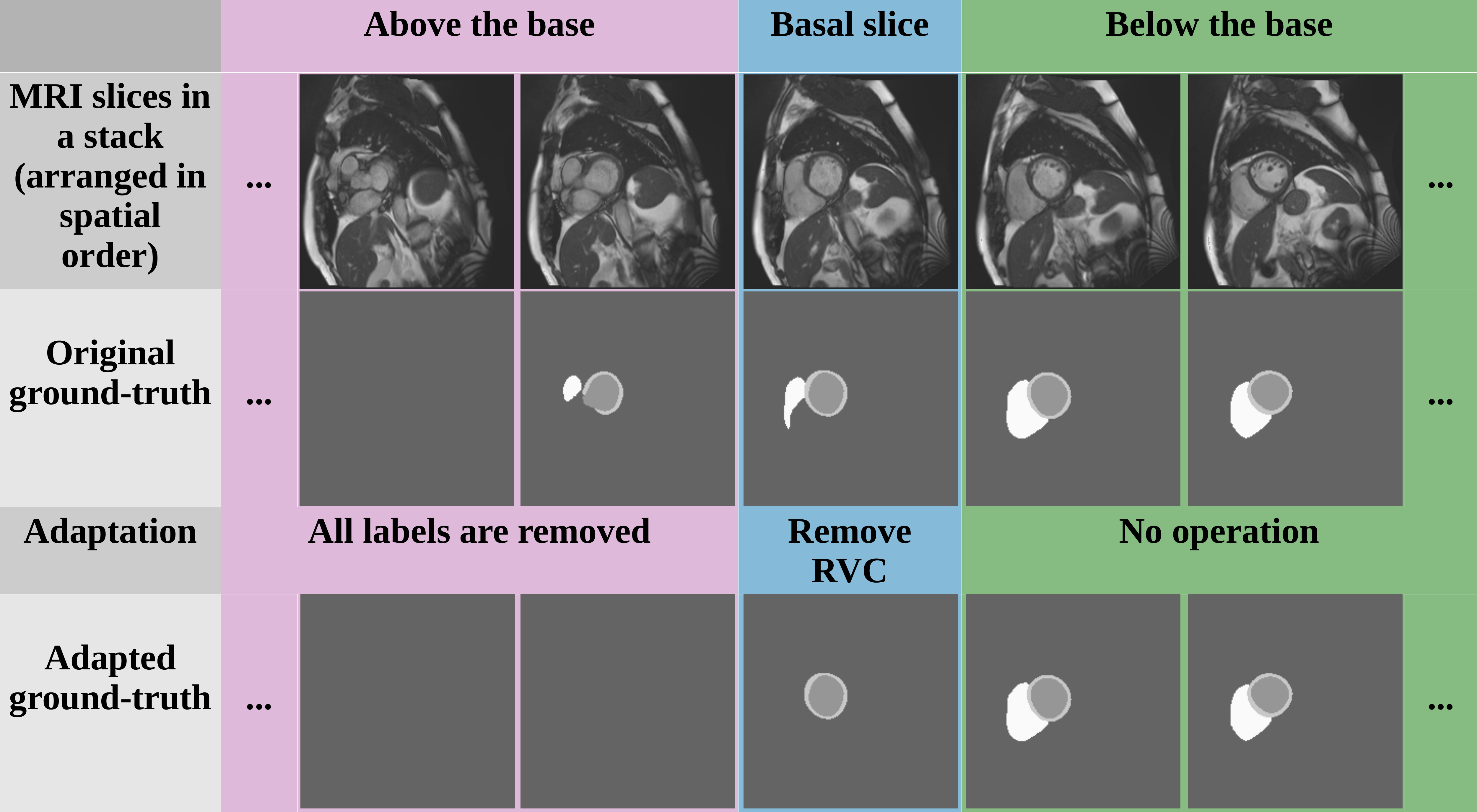}}
\caption{(Left) Intra- and inter-dataset inconsistencies of the basal slice ground-truth (RVSC contains no basal slice and is therefore not shown). \quad\quad  (Right) Ground-truth adaptation proposed for UK Biobank. the basal slice is first identified (blue), then the RVC labels are removed in this slice, and the labels are removed from the slices above (pink).}
\end{figure*}

\section{Data}

\subsection{Datasets}
The proposed method was trained using four datasets: the very large UK Biobank\cite{Petersen:2016} dataset through our access application \footnote{Application Number 2964.}, the ACDC challenge training dataset, the Sunnybrook dataset \cite{Radau:2009} (made available for the MICCAI 2009 challenge on automated left ventricle (LV) segmentation), and the Right Ventricle Segmentation Challenge (RVSC) dataset \cite{Petitjean:2015} (provided for the MICCAI 2012 challenge on automated right ventricle (RV) segmentation). Depending on the dataset, expert manual segmentation for different cardiac structures (e.g. the left and right ventricular cavity (LVC, RVC), the left ventricular myocardium (LVM)) is provided as ground-truth for all slices at end-diastole (ED) and/or end-systole (ES) phases. All other structures in the image are considered as background (BG). Training involved a subset (80\%) of the UK Biobank dataset. Testing used the remaining 20\% from the same dataset, as well as the whole three other datasets. Details about these datasets are provided in Appendix A. We mainly adopt the metrics used in the three challenges above to measure segmentation performance. The exact definitions of the metrics used in this paper (e.g. Dice index, Hausdorff distance, presence rate) are provided in Appendix B.

\subsection{Notation and Terminology}
In this paper, slices in image stacks are indexed in spatial order from the basal to the apical part of the heart.  Given an image stack $S$, we denote $N$ the number of its slices. Given two values $a$ and $b$ between $0$ and $N$, we note $S[a,b]$ the sub-stack consisting of slices of indexes in the interval $[round(a), round(b)[$ ($round(a)$ is included while $round(b)$ is excluded) with $round$ the function rounding to nearest integer. For instance, if $S$ is a stack of $N$=$10$ slices of indexes from 0 to 9, then $S[0.2N, 0.6N]$ is the stack consisting of slices number 2 to 5. Similarly, if the basal slice is defined in $S$, we denote $base$ its index. Then $S[base]$ and $S[base$+$1]$ are the basal slice and the first slice below the base.

Segmentation of slices above and below the base of the heart can be quite different. For convenience, in a stack with known base slice, we call the slices located above it the AB (above-the-base) slices, and the ones located below it BB (below-the-base) slices. In the remainder of this paper, we propose methods to determine the base slice for image stacks of UK Biobank using the provided ground-truth.

Finally, given a segmentation mask $M$, $edge(LVC, LVM)$ is the number of pairs of neighboring pixels (two pixels sharing an edge, defined using the 4-connectivity) on $M$ such that one is labeled to LVM while the other is to LVC. Similarly we define $edge(LVC, BG)$ and $edge(LVC, RVC)$.

\subsection{Adaptation of the UK Biobank Ground-Truth}

Let's first compare the segmentation conventions followed by the ground-truth between datasets. For BB slices, the convention is roughly the same: if LV is segmented, LVC is well enclosed in LVM; if RVC is segmented, it is identified as the whole cardiac cavity zone next to the LV. But for AB slices, the variability of segmentation conventions within and between datasets can be significant. On the left of Fig.1, we show examples of (base slice, ground-truth) pairs from UK Biobank (row-1 and row-2, two different cases), ACDC (row-3) and Sunnybrook (row-4). For better visualization, we crop out the heart regions from the original MRI images and ground-truths accordingly. The segmentation ground-truth on these similar images are significantly different. In particular, we notice the intra- and inter-dataset inconsistencies in the segmentation of (1) the RVC at the outflow tract level, (2) the LVM and LVC at the mitral valve level (some dataset seems to be segmented in a way such that the LVC mask is always fully surrounded by the LVM mask). In contrast, the convention seems roughly the same for the BB slices.

Hence we decided to adopt the ground-truth of UK Biobank to improve both consistency and generality. As presented in the right part of Fig.1, we i) set all pixels in all the slices above the base to BG; ii) relabel all the pixels in the basal slice originally labeled as RVC to BG while keeping the LVC and LVM pixels unchanged; iii) keep the ground-truth of all slices below the base unchanged. 

Moreover, we propose a method to determine the basal slice automatically in the stacks of UK Biobank. While checking the ground-truth of the slices starting from the apex part, the basal slice is determined as the first one such that:\\
- the LVC mask is not fully surrounded by the LVM mask:
\begin{equation}
edge(LVC,BG)+edge(LVC,RVC)>0
\end{equation}
- or the area of the RVC mask shrinks substantially comparing to that of the slice below:
\begin{empheq}[left=\empheqlbrace]{align}
&overlap(RVC_1, RVC_2) / RVC_2 \leq T_1\\
&RVC_1 / RVC_2 \leq T_2
\end{empheq} 
with $RVC_1$ and $RVC_2$ the RVC masks of the slice and the slice below it respectively,  $T_1$=$0.75$ and $T_2$=$0.8$ thresholds. 
If the basal slice is not determined after examining all slices in the stack, we define that the index of the base slice is $-1$ (so $S[base$+$1]$ is the first slice in the stack).

According to the current international guidelines in \cite{Schulz-Menger:2013}, the ``standard" basal slice is the topmost short-axis view slice that has more than 50\% myocardium around the blood cavity. To test whether the UK Biobank basal slices determined above are close to the standard basal slices, we randomly picked 50 cases (50 ED stacks + 50 ES stacks) and estimated their standard basal slices at ED and ES visually according to the guidelines. Among the 100 pairs of standard basal slice and ground-truth-deduced basal slice, 59 pairs are exactly the same, 40 pairs are 1-slice away in stack, and only 1 pair is 2-slice away. The ``adapted" ground-truth will stand as the ground-truth for the rest of this paper.

\section{Methods}

\begin{figure*}[]
\centering
\subfigure{
\includegraphics[width=5.7cm, height=7.5cm]{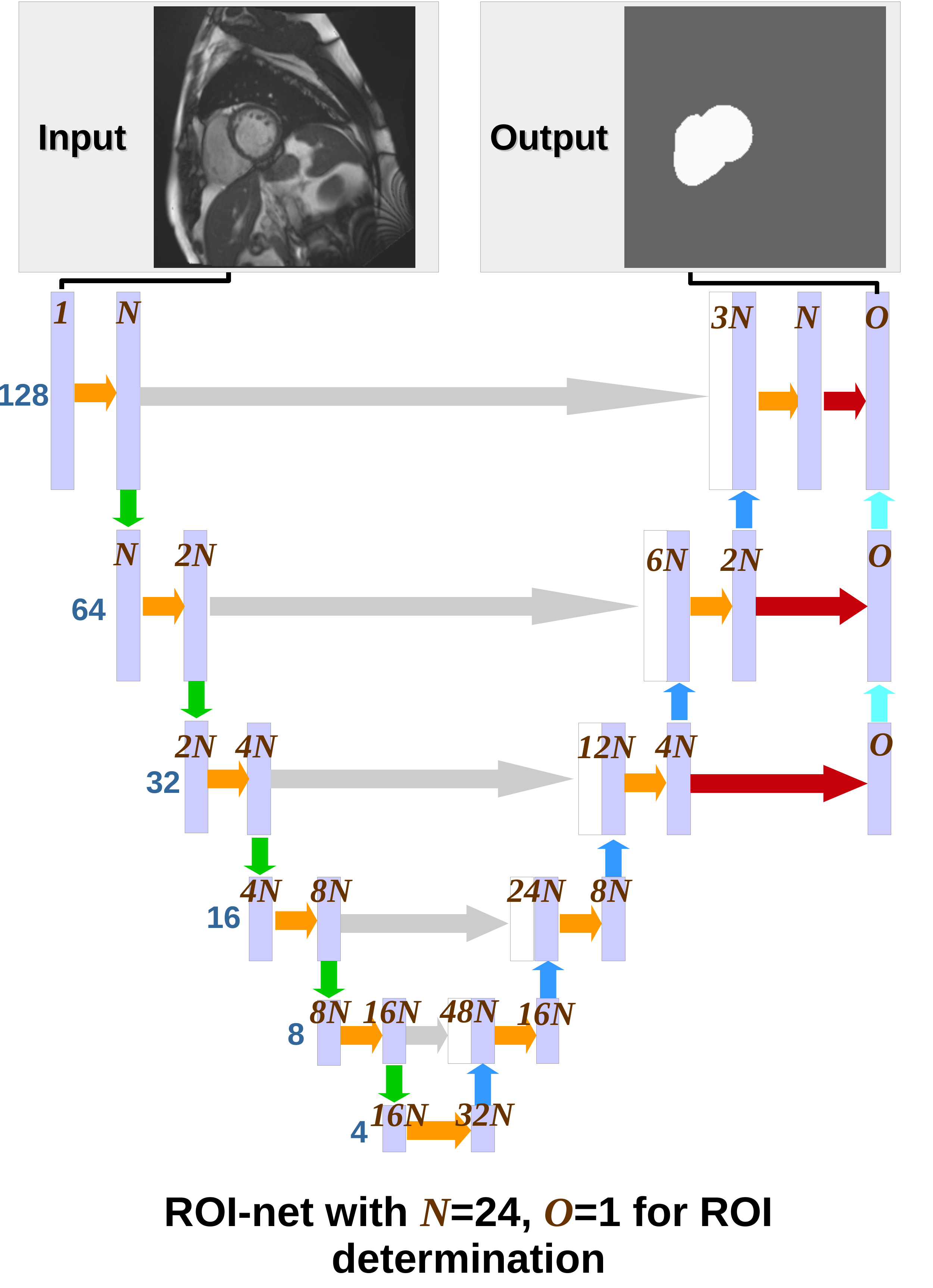}}\quad
\subfigure{
\includegraphics[width=9.8cm, height=7.5cm]{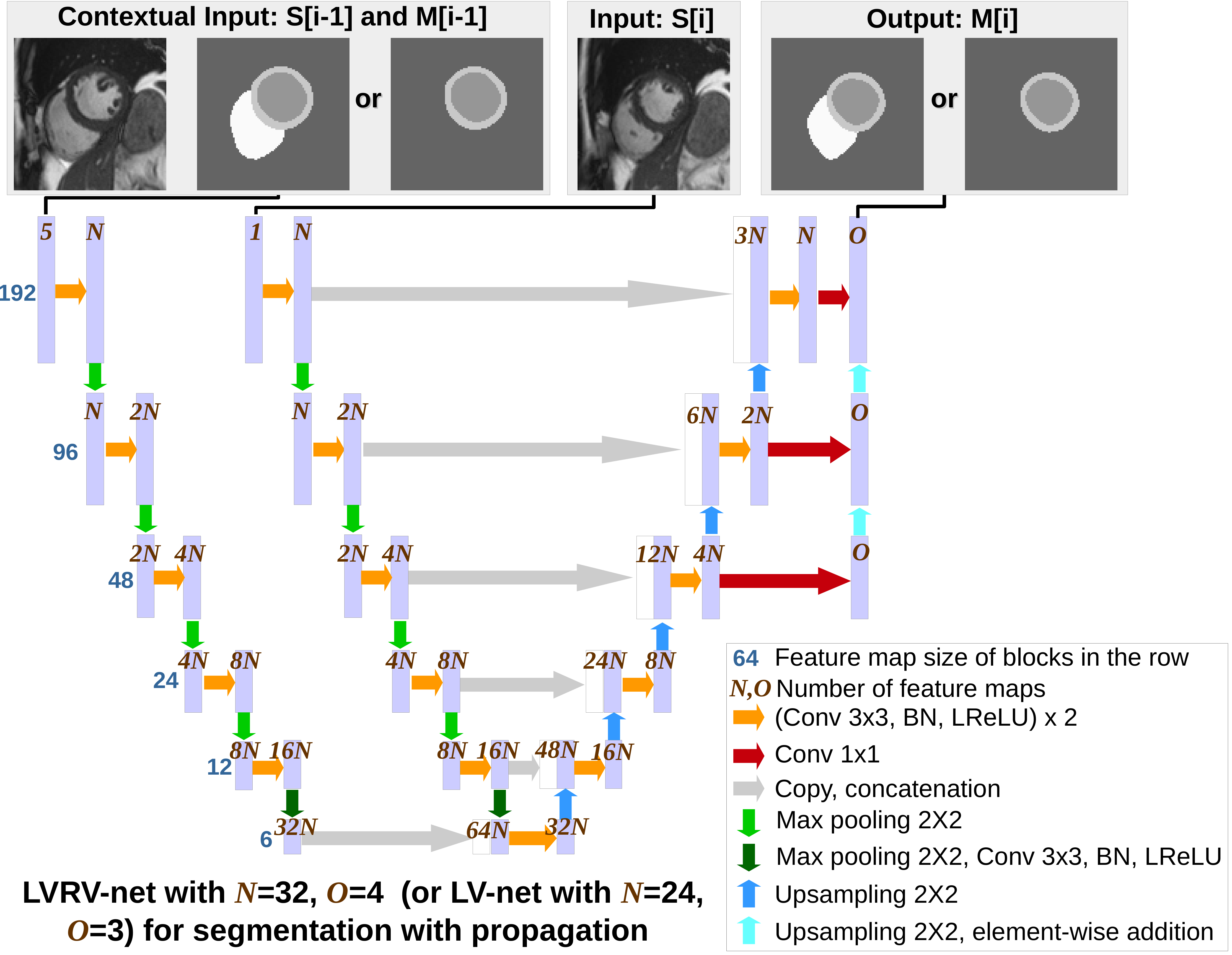}}
\caption{(Left) ROI-net: for ROI determination over image stack. A sigmoid function is applied to the output channel to generate pixel-wise probabilities. \quad  (Right) LVRV-net and LV-net: for cardiac segmentation on ROIs. S[i] is the slice to be segmented and M[i] is the predicted mask. A softmax function is applied to the output channels to generate pixel-wise 4- or 3-class probabilities.}
\end{figure*}

Our method mainly consists of two steps: region of interest (ROI) determination and segmentation with propagation. The first step is either based on a trained neural network (the ROI-net) or on center cropping, depending on the dataset. The second step is based on either the LVRV-net or the LV-net (originally designed by us and inspired from U-net \cite{Ronneberger:2015}), depending on whether the RVC must be segmented. This section will also present the image preprocessing methods and the loss functions we used.

\subsection{Region of Interest (ROI) Determination: ROI-net}

On cardiac MRI images, defining an ROI is useful to save memory usage and to increase the speed of segmentation methods. There are many different ROI determination methods available in the community. But for most of them, the robustness remains a question, as the training and the evaluation are done with cases from the same cohort of limited size. We propose a robust approach as follows. With a large number of available cases from UK Biobank, a deep learning based method becomes a natural choice. In particular, we design the ROI-net to perform heart segmentation on MRI images.

Notice that for some datasets (Sunnybrook and RVSC), the images are already centered around the heart. Similar to what was done in \cite{Tran:2016}, in such cases, images are simply cropped. However this is not valid for most datasets (here UK Biobank and ACDC), and an ROI needs to be determined specifically for each stack based on the predictions of ROI-net, as explained in the following. ROI-net is a variant of U-net with a combination of convolutions, batch normalizations (BN) and leaky ReLUs (LReLU) \cite{Maas:2013} as building blocks. In leaky ReLU the gradient parameter when the unit is not active is set to $0.1$.  Furthermore, we implement deep supervision as in \cite{Kayalibay:2017} to generate low resolution (of size 32 and 64) segmentation outputs, and then upsample and add them to the final segmentation. A sigmoid function is applied to the output channel of ROI-net to generate pixel-wise probabilities. 

In brief, ROI-net takes one original MRI image as input and predicts pixel-wise probabilities as a way of heart/background binary segmentation (0 for background, 1 for the heart, and the threshold is 0.5 in inference). The heart to be segmented is defined as the union of LVC, LVM, and RVC. The ROI determination takes only the ED stack slices into account. In practice, an ROI containing the heart with some margin at ED also contains well the heart at other instants including ES. More specifically:
\subsubsection{Training}
The network is trained with slices in $S[(base$+$1), (base$+$1)$+$0.4N]$ (the 40\% of slices just below the base) of the ED stack $S$ from the UK Biobank training cases. The purpose of using only slices in $S[(base$+$1), (base$+$1)$+$0.4N]$ is to avoid the top slices around the base on which RVC ground-truth shrinks (Fig.1), and the bottom slices around the apex on which the heart is small and almost does not affect the ROI determination.
\subsubsection{Prediction}
To confirm the robustness of ROI-net for inference, we apply it to the sub-stacks roughly covering the largest cross-section of the hearts in a dataset (the position of the base is supposed to be unknown for individual cases). The slice indexes of these sub-stacks are determined based on visual observation for a given dataset. More specifically, the trained ROI-net is used to segment slices in $S[0.2N, 0.6N]$ of the ED stack $S$ of all the UK Biobank cases, and slices in $S[0.1N, 0.5N]$ of the ED stack $S$ of all the ACDC cases. For noise reduction and as post-processing for the ROI net, for each image, only the largest connected component of the output heart mask is kept for prediction.
\subsubsection{ROI Determination}
For each ED stack, the union of all predicted heart masks, as well as the minimum square $M$ covering their union, is determined. We add to it a padding of width $0.3$ times the size of $M$ to generate a larger square bounding box, which is defined as the ROI for the case.

After ROI determination on an ED stack, the same ROI applies to both the ED and ES stacks of the same case. Then the ED and ES stacks are cropped out according to this ROI and used as inputs for the LVRV-net and the LV-net in the second step. Hence in the remainder of this paper, we refer to the cropped version of the images, slices or stacks.

\subsection{Segmentation with Propagation: LVRV-net and LV-net}
The second step is segmenting the cropped images (the ROIs). Depending on whether we segment RVC or not, we proposed two networks: LVRV-net and LV-net. They share the same structure template as depicted on the right of Fig.2. Both perform slice segmentation of $S[i]$ taking $S[i$-$1]$, the adjacent slice above, and $M[i$-$1]$ its segmentation mask, as contextual input. In the contextual input, there are five channels in total: $S[i$-$1]$ takes one, while $M[i$-$1]$, being converted to pixel-wise one-hot channels (BG, LVC, LVM, RVC), takes four. In case $S[i]$ is the first slice to be segmented in a stack, $M[i$-$1]$ does not exist and is set to a null mask; in case $S[i]$ is the top slice in a stack, $S[i$-$1]$ does not exists and is set to a null image. The main body of LVRV-net and LV-net is also a variant of U-net with convolution, BN, LReLU and deep supervision, very similar to that of ROI-net. In addition to the main body, an extra encoding branch encodes the contextual input. Information extracted by the main body encoding branch and the extra encoding branch are combined at the bottom of the network, before being decoded in the decoding branch. Finally, a softmax function is applied to the output channels to generate pixel-wise 4- or 3-class probabilities. For inference, each pixel is labeled to the class with the highest probability.

\subsubsection{Training}
LVRV-net and LV-net are trained to segment slices $S[i]$ in $S[(base$+$1),N]$ (the BB slices, the green column in Fig.1) and $S[base,N]$ (the basal slice and the BB slices, the blue column and the green columns in Fig.1) respectively  of the stack $S$ at ED and ES of the UK Biobank training set. Regarding the contextual input, $S[i$-$1]$ and $M[i$-$1]$ are set to a null image or a null mask if they are not available as described above; otherwise $M[i$-$1]$ is set to the corresponding ground-truth mask.
\subsubsection{Testing}
The trained LVRV-net and LV-net are used to segment the cases in the UK Biobank testing set and the other datasets (ACDC, Sunnybrook, RVSC). Let us note $S'$ the sub-stack to be segmented and $M'$ the corresponding predicted mask stack. Notice that for UK Biobank, $S'$ is $S[(base$+$1),N]$ for LVRV-net, and $S[base,N]$ for LV-net; for the other datasets, $S'$ is the whole stack. LVRV-net or LV-net iteratively segments $S'[i]$ by predicting $M'[i]$, taking $S'[i$-$1]$ and $M'[i$-$1]$ as contextual input, for $i$ = 0, 1, 2, etc.. In other words, the segmentation prediction of a slice is used as contextual information while segmenting the slice immediately below it in the next iteration. The segmentation prediction is iteratively ``propagated" from top to bottom (or roughly speaking from base to apex) slices in $S'$. 
\subsubsection{Post-processing}
We post-process the predictions at each iteration while segmenting a stack (hence the post-processed mask will be used as the contextual mask in the next iteration if it exists). A predicted mask is considered as successful if the two conditions below are satisfied:\\
- LVM is present on the mask;\\
- LVC is mostly surrounded by LVM:
\begin{equation}
\begin{split}
\big( \hspace{0.5ex} edge(LVC, BG) + edge(LVC, RVC) \hspace{0.5ex} \big) \\ 
\leq 0.5\times edge(LVC, LVM)
\end{split}
\end{equation}
The parameter $0.5$ above is determined empirically. If the predicted mask is successful, for LVRV-net only, we further process the mask by preserving only the largest connected component of RVC and turning all the other RVC connected components (if they exist) to background; otherwise, the predicted mask is reset to a null mask.

\subsection{Image Preprocessing}
Each input image or mask of the three networks in this paper (ROI-net, LVRV-net, and LV-net) is preprocessed as follows:

\subsubsection{Extreme Pixel Value Cutting and Contrast Limited Adaptive Histogram Equalization (CLAHE) for ROI-net only}
Input images to ROI-net are thresholded to the 5th and 95th percentiles of gray levels. Then we apply CLAHE as implemented in OpenCV \footnote{\url{https://docs.opencv.org/3.1.0/d5/daf/tutorial_py_histogram_equalization.html}} to perform histogram equalization and improve the contrast of the image with the parameters $clipLimit=3$ and $tileGridSize=(8,8)$. 
\subsubsection{Padding to Square and Resize}
The input image or mask is zero-padded to a square if needed. Then it is resampled using nearest-neighbor interpolation to $128\times128$ for ROI-net or $192\times192$ for LVRV-net and LV-net.
\subsubsection{Normalization}
Finally, for each input image of all networks, the mean and standard deviation of the slice intensity histogram cropped between the 5th and 95th percentiles are computed. The image is then normalized by subtracting this mean and dividing by this standard deviation.

\subsection{Loss Functions}
We use the two Dice loss (DL) functions below to train the three neural networks mentioned above. As suggested in \cite{Wolterink:2017}, loss functions based on Dice index help overcoming difficulties in training caused by class imbalance.

\subsubsection{$DL_1$ for ROI-net Training}
Given an input image $I$ of $N$ pixels, let's note $p_n$ the pixel-wise probability predicted by ROI-net and $g_n$ the pixel-wise ground-truth value ($g_n$ is either $0$ or $1$). $DL_1$ is defined as
\begin{equation}
DL_1 = -\frac{ 2\sum_{n=1}^{N} p_n g_n + \epsilon}{ \sum_{n=1}^{N} p_n + \sum_{n=1}^{N} g_n + \epsilon}
\end{equation}
where $\epsilon$ is used to improve the training stability by avoiding division by 0, i.e. when $p_n$ and $g_n$ are 0 for each pixel $n$. Empirically we take $\epsilon = 1$. The value of DL1 varies between 0 and -1. Good performance of ROI-net corresponds to $DL_1$ close to -1.

\subsubsection{$DL_2$ for LVRV-net Training}
For the segmentation of a $N$-pixel input image, the outputs are four probabilities $p_{n,c}$ with $c = 0, 1, 2, 3$ (BG, LVC, LVM and RVC) such that $\sum_{c} p_{n,c} = 1$ for each pixel. Let's note $g_{n,c}$ the corresponding one-hot ground-truth ($g_{n,c}$ is 1 if the pixel is labeled with the class corresponding to $c$; otherwise $g_{n,c}$ is 0). Then we define
\begin{equation}
DL_2 = - \frac{1}{4} \sum_{c=0}^{3} ( \frac{ 2\sum_{n=1}^{N} p_{n,c} g_{n,c} + \epsilon}{ \sum_{n=1}^{N} p_{n,c} + \sum_{n=1}^{N} g_{n,c} + \epsilon} )
\end{equation}
The role of $\epsilon$ here is similar to that in $DL_1$. Empirically we use $\epsilon = 1$.

\subsubsection{$DL_3$ for LV-net Training}
Its formula is very similar to that of $DL_2$. The only difference is, instead of calculating the average of the 4 Dice index terms with $c$ ranges from 0 to 3, $DL_3$ sums up the 3 Dice index terms with $c$ ranges from 0 to 2 (BG, LVC, LVM) and computes their average.


\section{Experiments and Results}
The three networks (ROI-net, LVRV-net, LV-net) are implemented using TensorFlow \footnote{\url{https://www.tensorflow.org/}} and trained with 3078 UK Biobank cases as described in the ``Methods" section. Then they are applied to the other datasets (ACDC, Sunnybrook, RVSC) without any fine-tuning or further training.

\subsection{Technical Details about Training the Three Networks}
ROI-net is trained for 50 epochs and applied to these cases to determine the ROIs. The cropped ROI volumes are then used to train LVRV-net and LV-net for 80 epochs. For each of the three networks, weights are initialized randomly, Adam optimizer is used with initial learning rate 0.0001, batch size is set to 16, and data augmentation is applied (the input images are randomly rotated, shifted and zoomed in/out along the row/column dimension independently, flipped horizontally and flipped vertically).

\subsection{Experiments on UK Biobank \& Contribution of the Propagation}

\begin{figure}[]
\centering
\includegraphics[width=8.4cm, height=2.1cm]{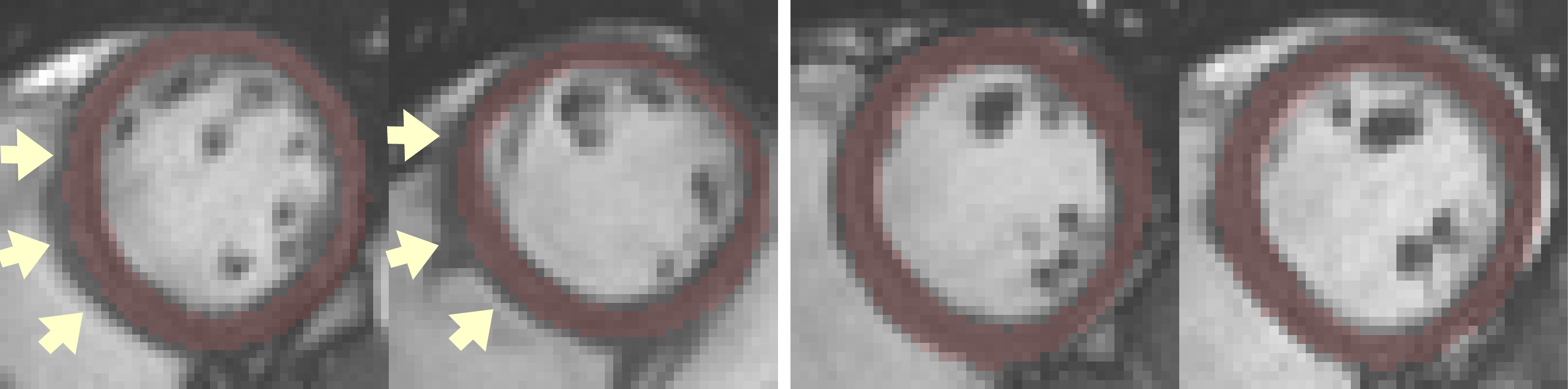}
\caption{UK Biobank ground-truth variability: These slices are extracted from 4 different cases in UK Biobank. Compared to the ground-truth of the 3rd and 4th slices, the ground-truth of the 1st and 2nd slices clearly under-segments the portion of myocardium between LV and RV (indicated by the arrows).}
\end{figure}

\begin{table*}
\caption{Segmentation Results on the UK Biobank Testing Cases}
\centering
\begin{tabular}{c cccccccc}
\hline
\noalign{\vskip 0.0in}
\multicolumn{1}{|}{} & \multicolumn{4}{|c|}{Dice} & \multicolumn{4}{c|}{Hausdorff (mm)} \\ 
\hline
\multicolumn{1}{|}{} & \multicolumn{1}{|c|}{LVM} & \multicolumn{1}{c|}{LVC} & \multicolumn{1}{c|}{LV-epi} & \multicolumn{1}{c|}{RVC} & \multicolumn{1}{c|}{LVM} & \multicolumn{1}{c|}{LVC} & \multicolumn{1}{c|}{LV-epi} & \multicolumn{1}{c|}{RVC}\\
\hline
\noalign{\vskip 0.0in}
\multicolumn{1}{|c}{proposed LVRV-net} & 0.769 (0.06) & 0.903 (0.03) & 0.932 (0.01) & 0.881 (0.04) & \textbf{7.66 (4.55)} & 5.94 (2.26) & \textbf{7.13 (4.32)} & \multicolumn{1}{c|}{10.39 (4.71)} \\
\multicolumn{1}{|c}{LVRV-mid-starting-net} & 0.767 (0.06) & 0.904 (0.03) & 0.931 (0.01) & 0.886 (0.04) & 8.96 (10.94) & \textbf{5.87 (2.90)} & 8.46 (10.98) & \multicolumn{1}{c|}{\textbf{9.90 (4.01)}}\\
\multicolumn{1}{|c}{LVRV-no-propagation-net} & \textbf{0.793 (0.05)} & \textbf{0.915 (0.03)} & \textbf{0.939 (0.01)} & \textbf{0.896 (0.03)} & 9.86 (12.03) & 6.66 (7.74) & 9.40 (12.03) & \multicolumn{1}{c|}{10.32 (5.32)}\\
\hline
\multicolumn{1}{|c}{proposed LV-net} & 0.752 (0.06) & 0.896 (0.04) & 0.923 (0.02) & - (-) & 9.78 (9.22) & 6.97 (3.43) & 8.72 (9.22) & \multicolumn{1}{c|}{- (-)}\\

\hline
\end{tabular}
\end{table*}

\begin{figure}[]
\centering
\includegraphics[width=3.6cm, height=3.2cm]{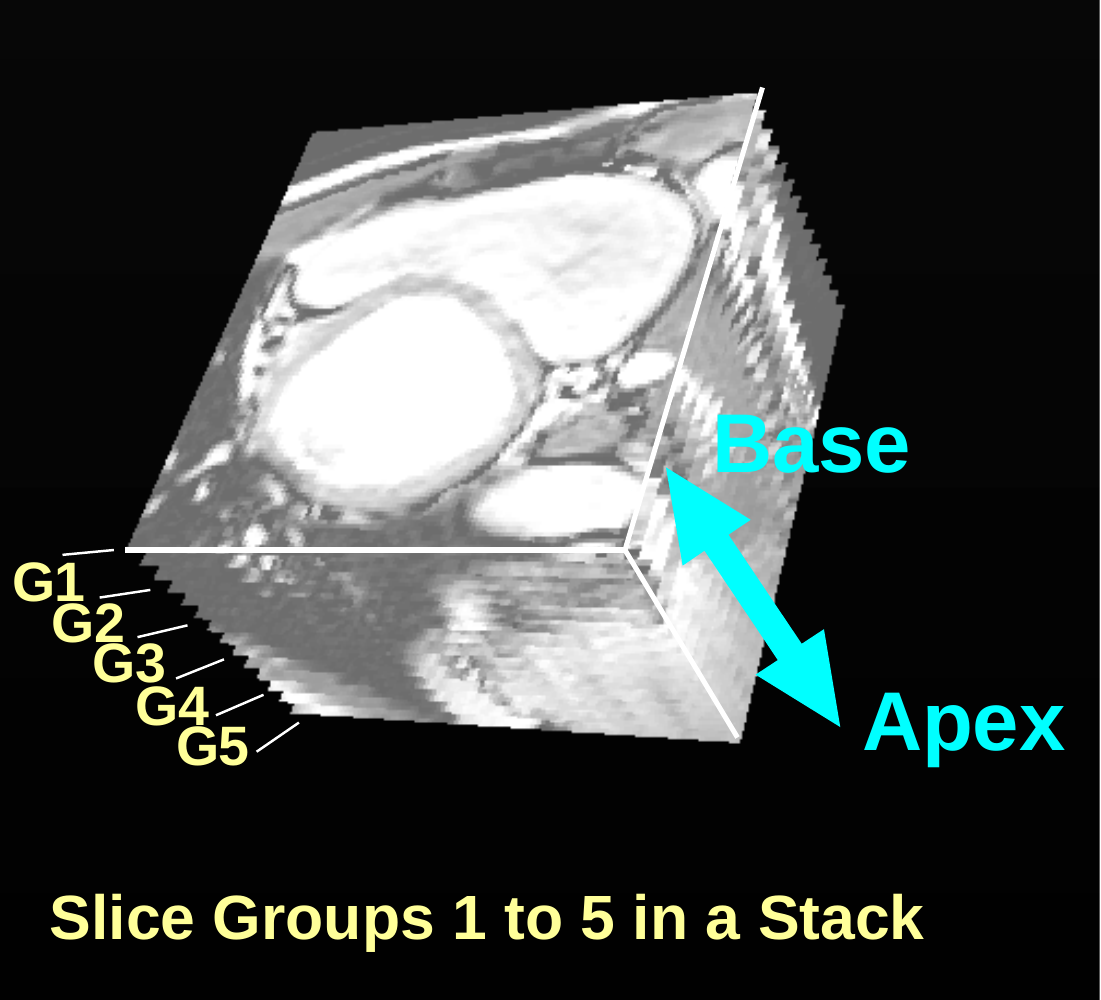}
\hfill
\includegraphics[width=4.6cm, height=3.2cm]{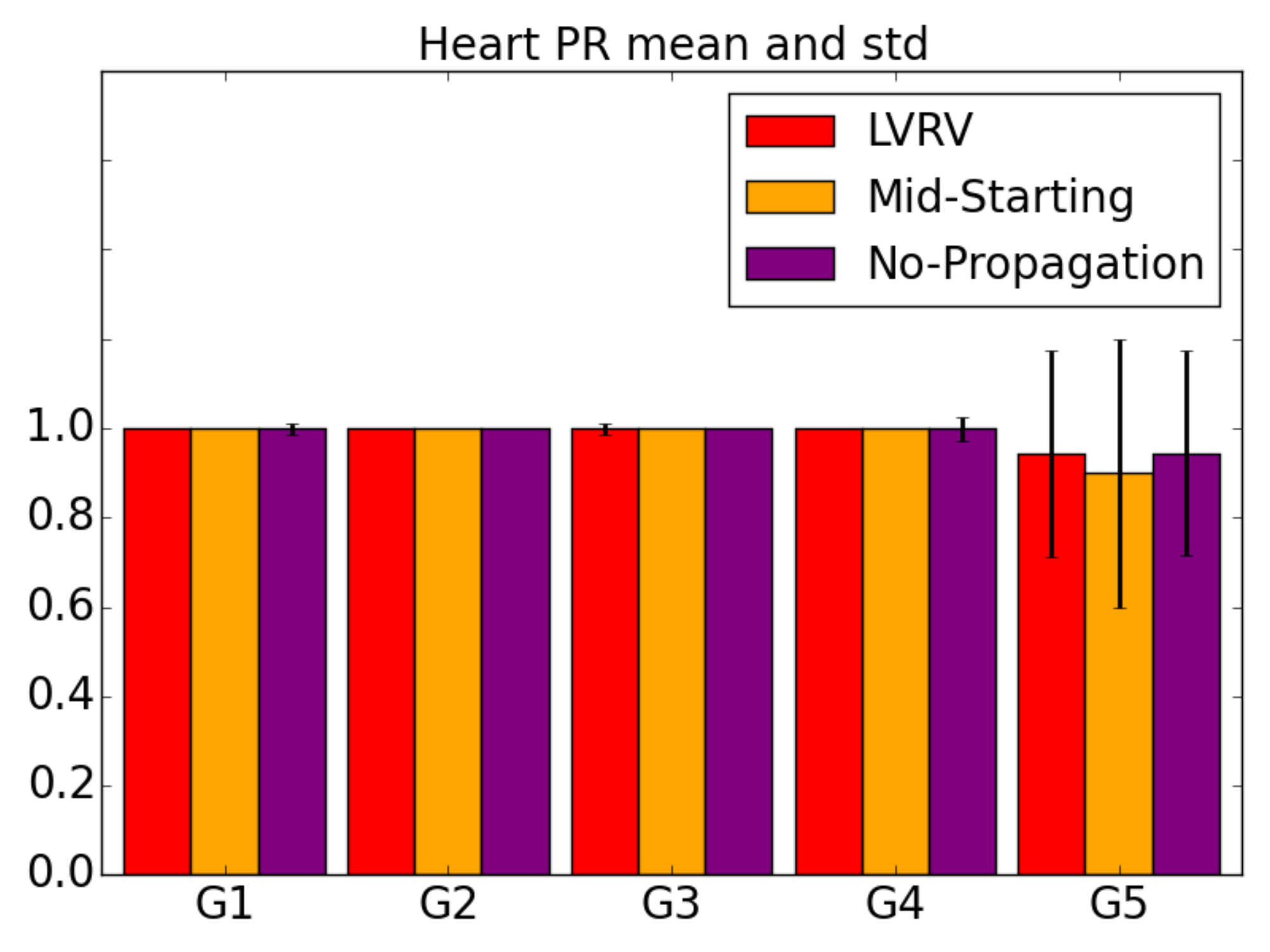}
\caption{(Left) An example of slice group division (G1 to G5) in a stack. \quad  (Right) Performance measured by heart presence rate (PR) of the LVRV-net, the LVRV-mid-starting-net and the LVRV-no-propagation-net on UK Biobank.}
\end{figure}

\begin{figure*}[]
\centering
\includegraphics[width=4.4cm, height=3.0cm]{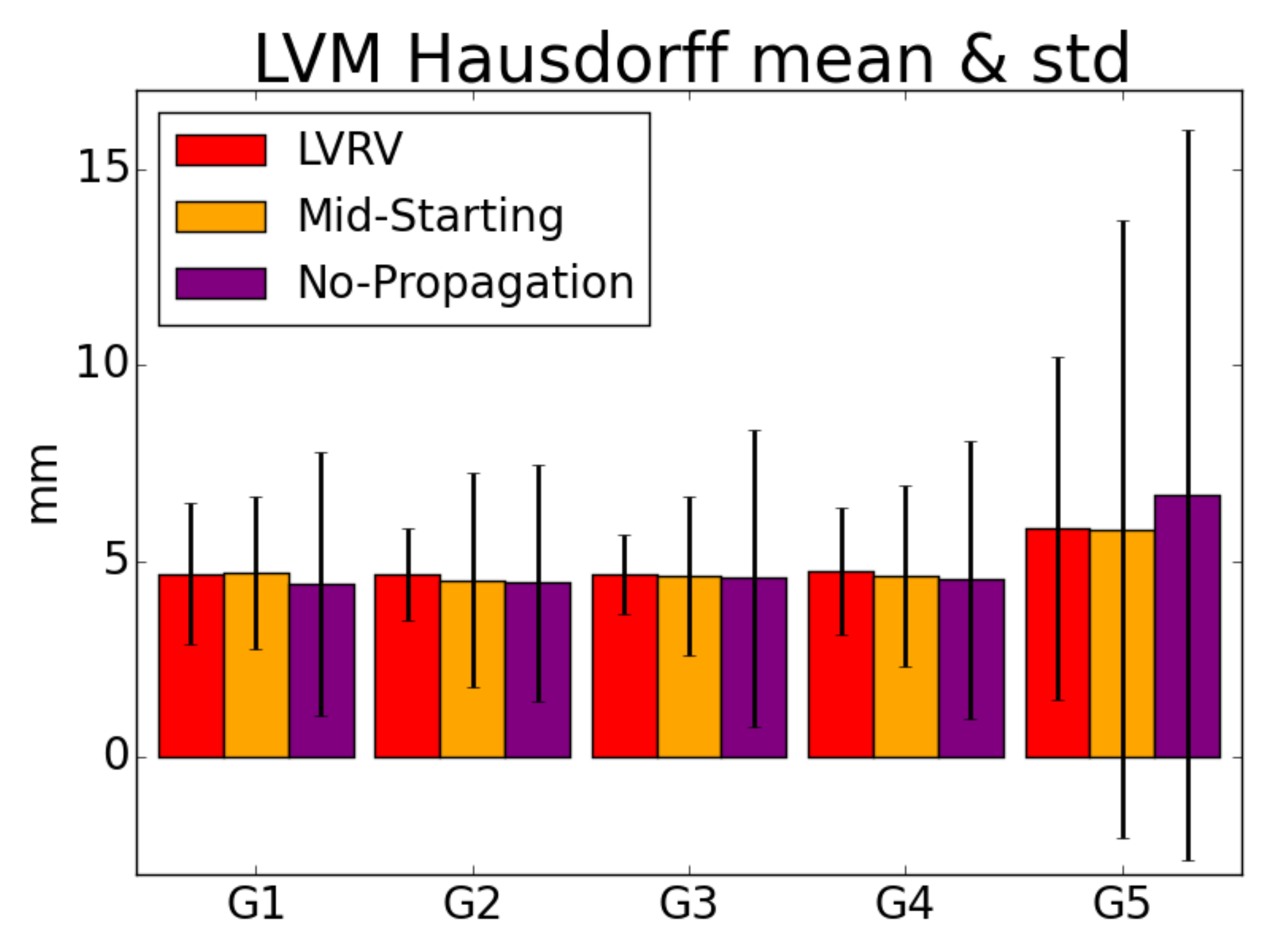}
\hfill
\includegraphics[width=4.4cm, height=3.0cm]{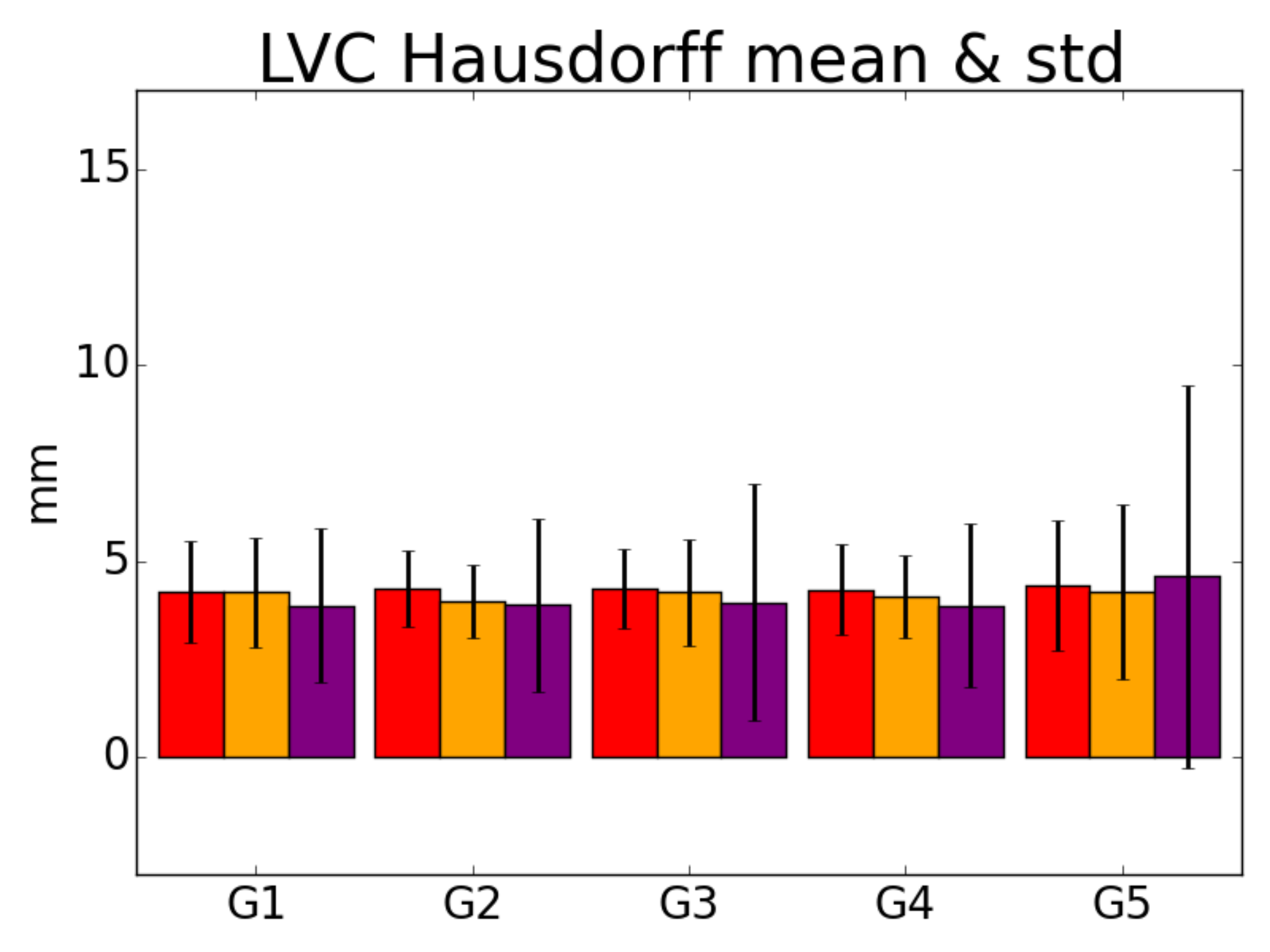}
\hfill
\includegraphics[width=4.4cm, height=3.0cm]{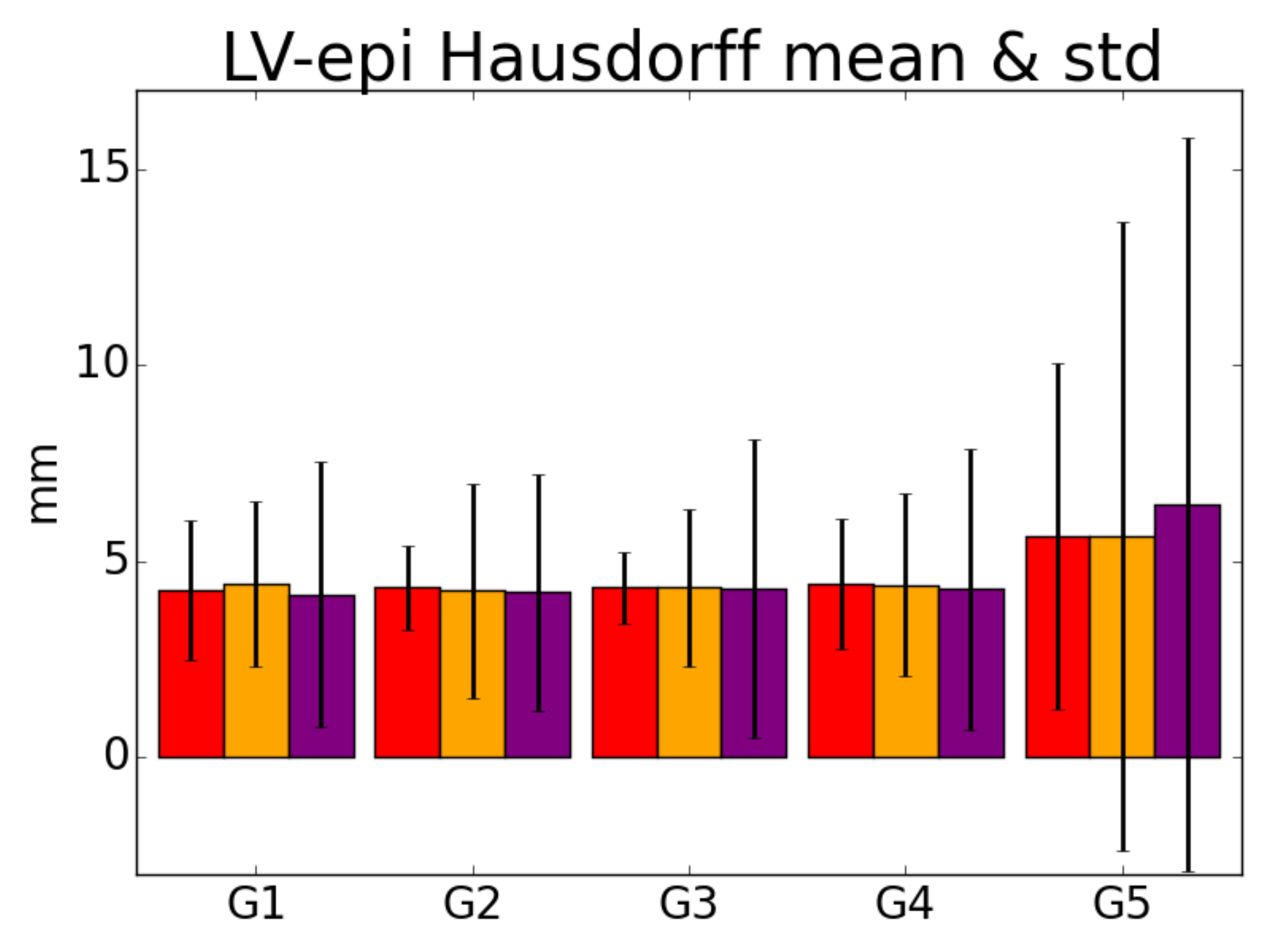}
\hfill
\includegraphics[width=4.4cm, height=3.0cm]{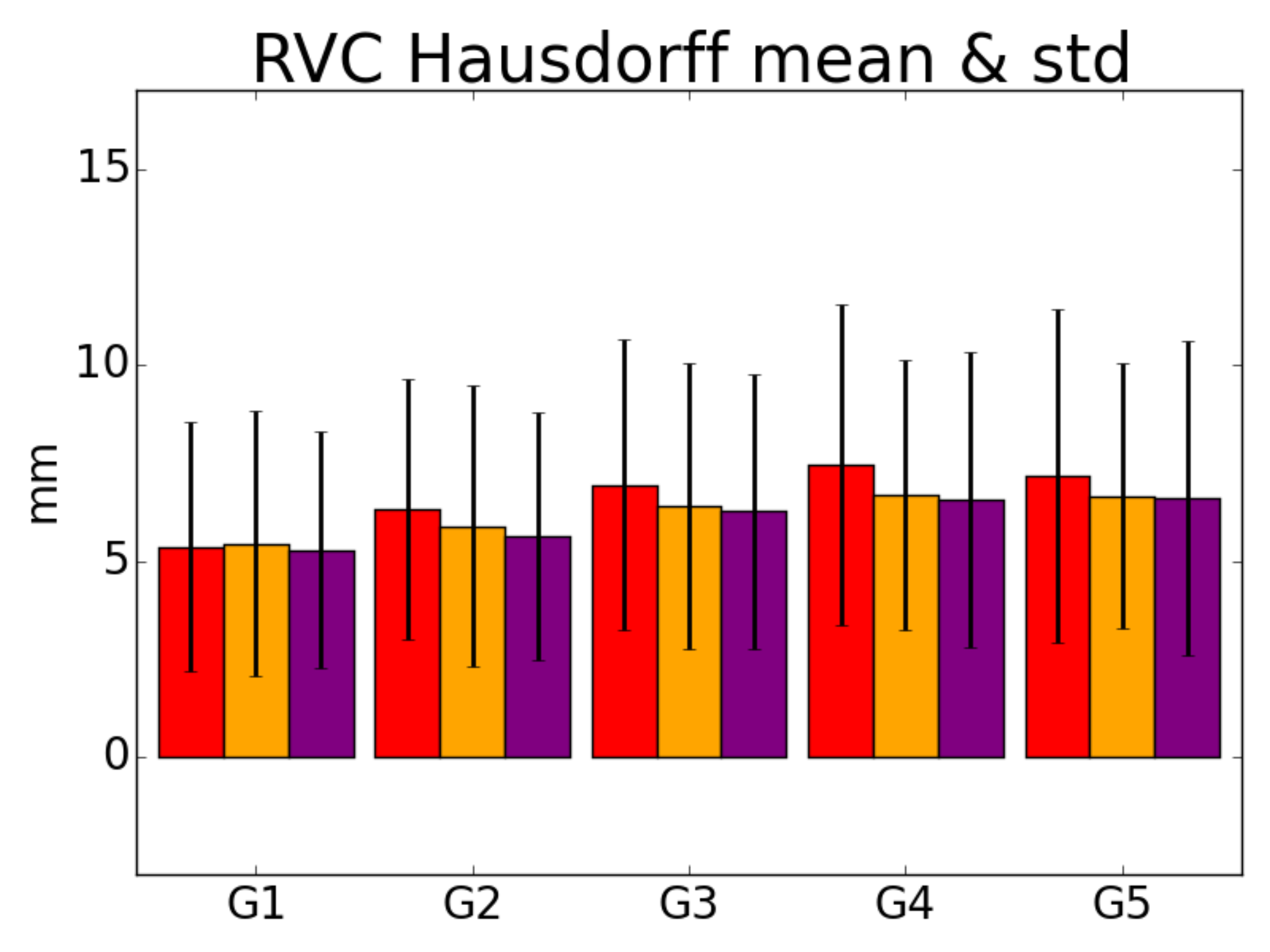}\\

\includegraphics[width=4.4cm, height=3.0cm]{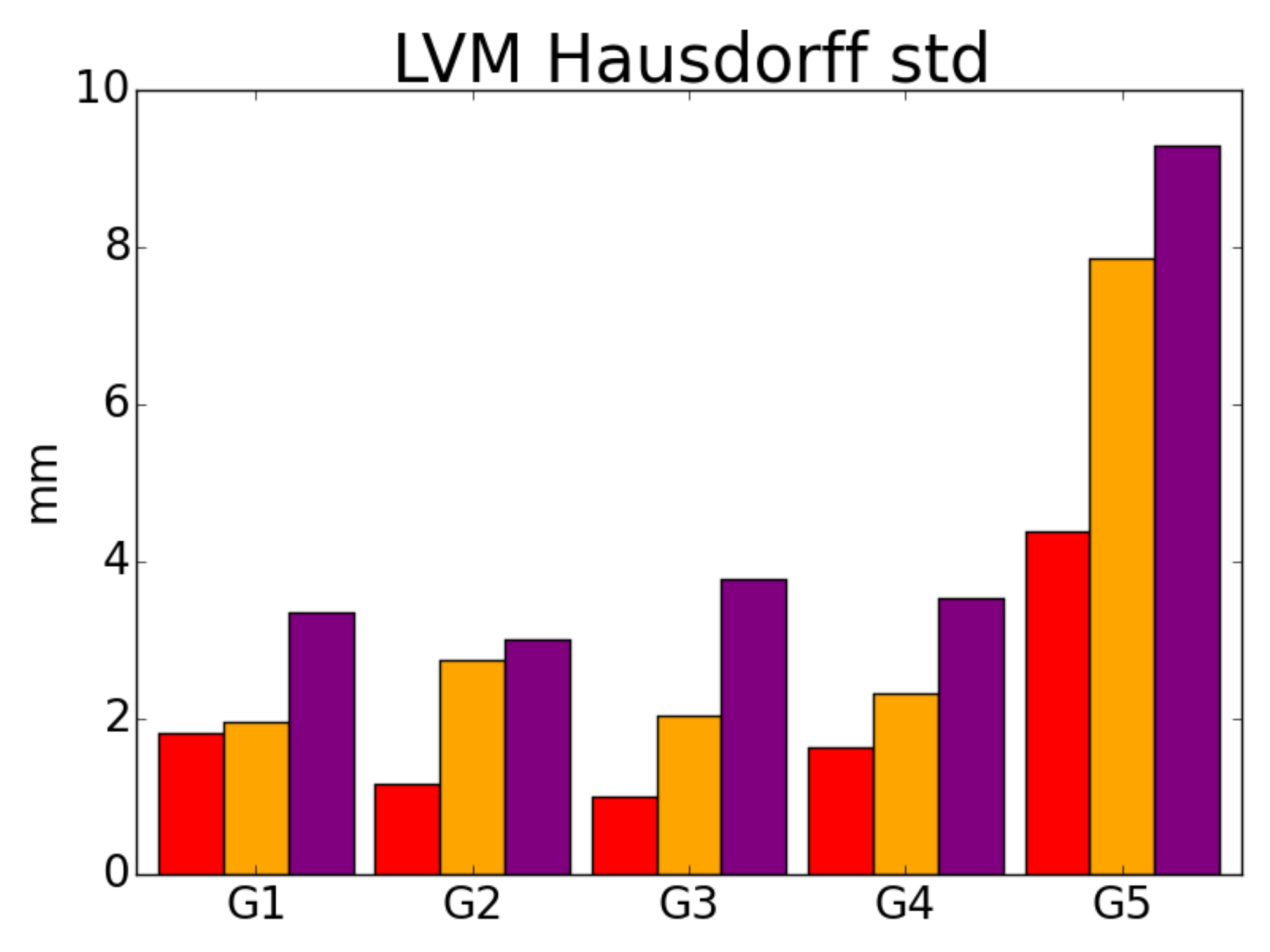}
\hfill
\includegraphics[width=4.4cm, height=3.0cm]{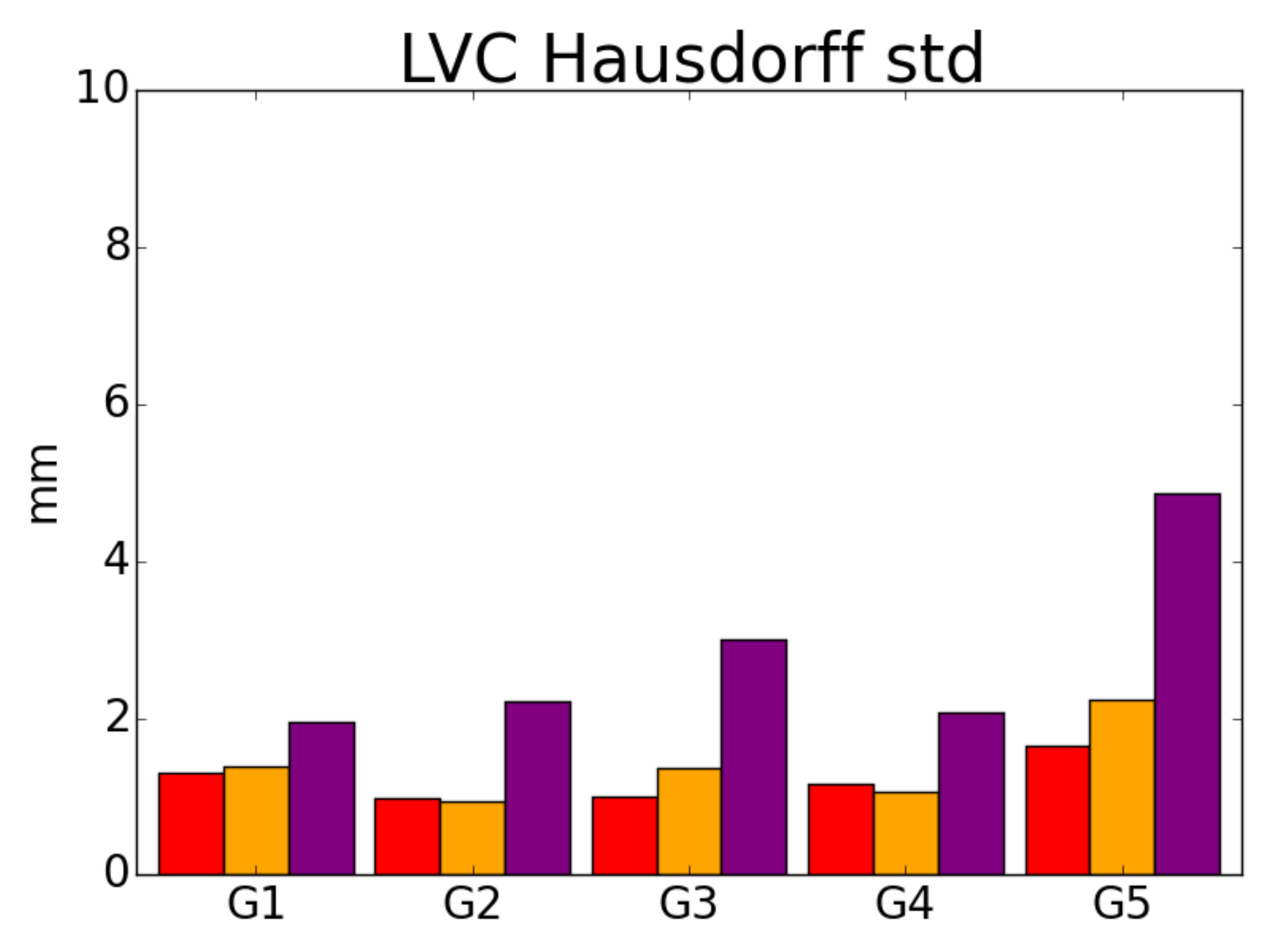}
\hfill
\includegraphics[width=4.4cm, height=3.0cm]{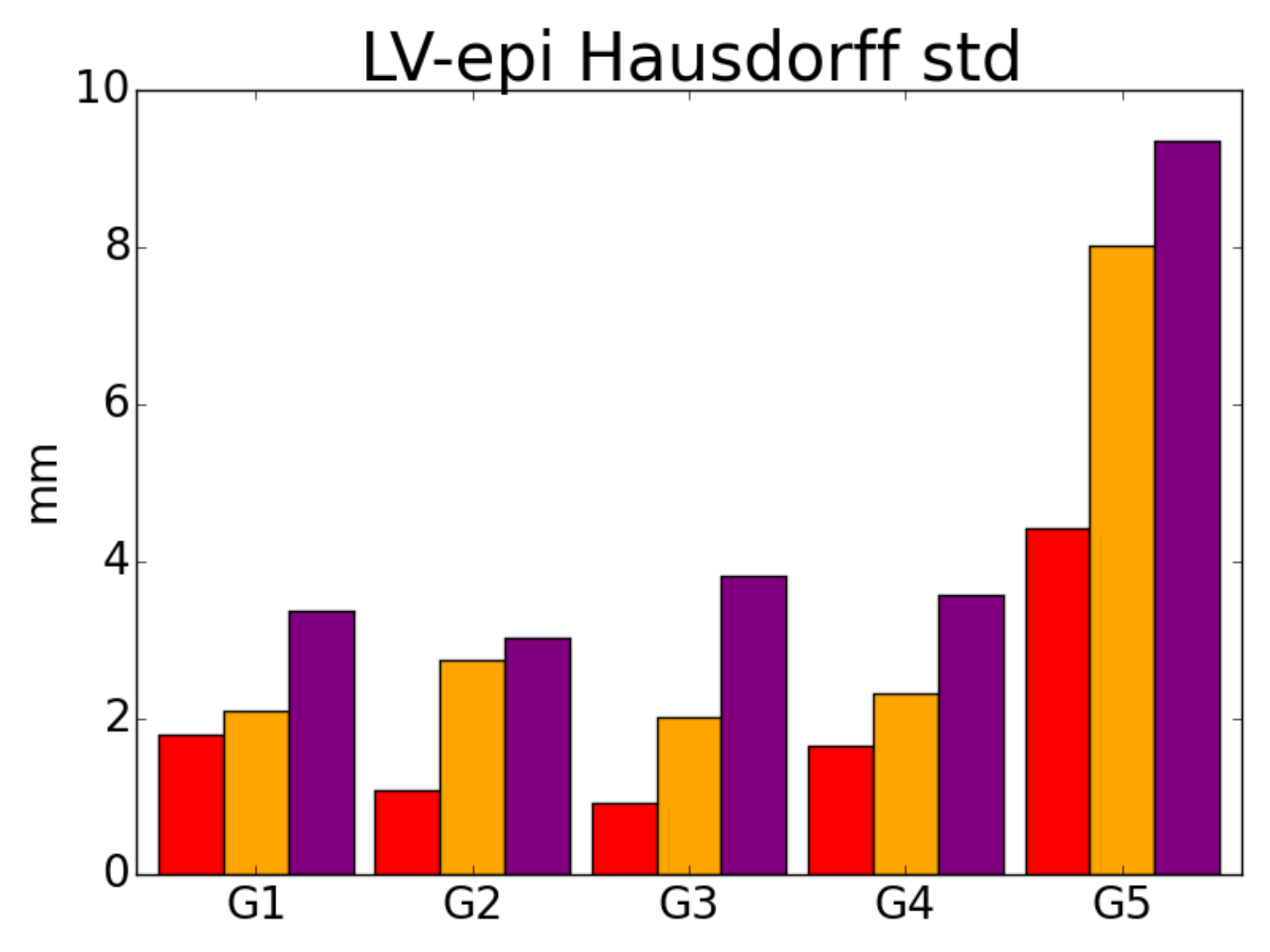}
\hfill
\includegraphics[width=4.4cm, height=3.0cm]{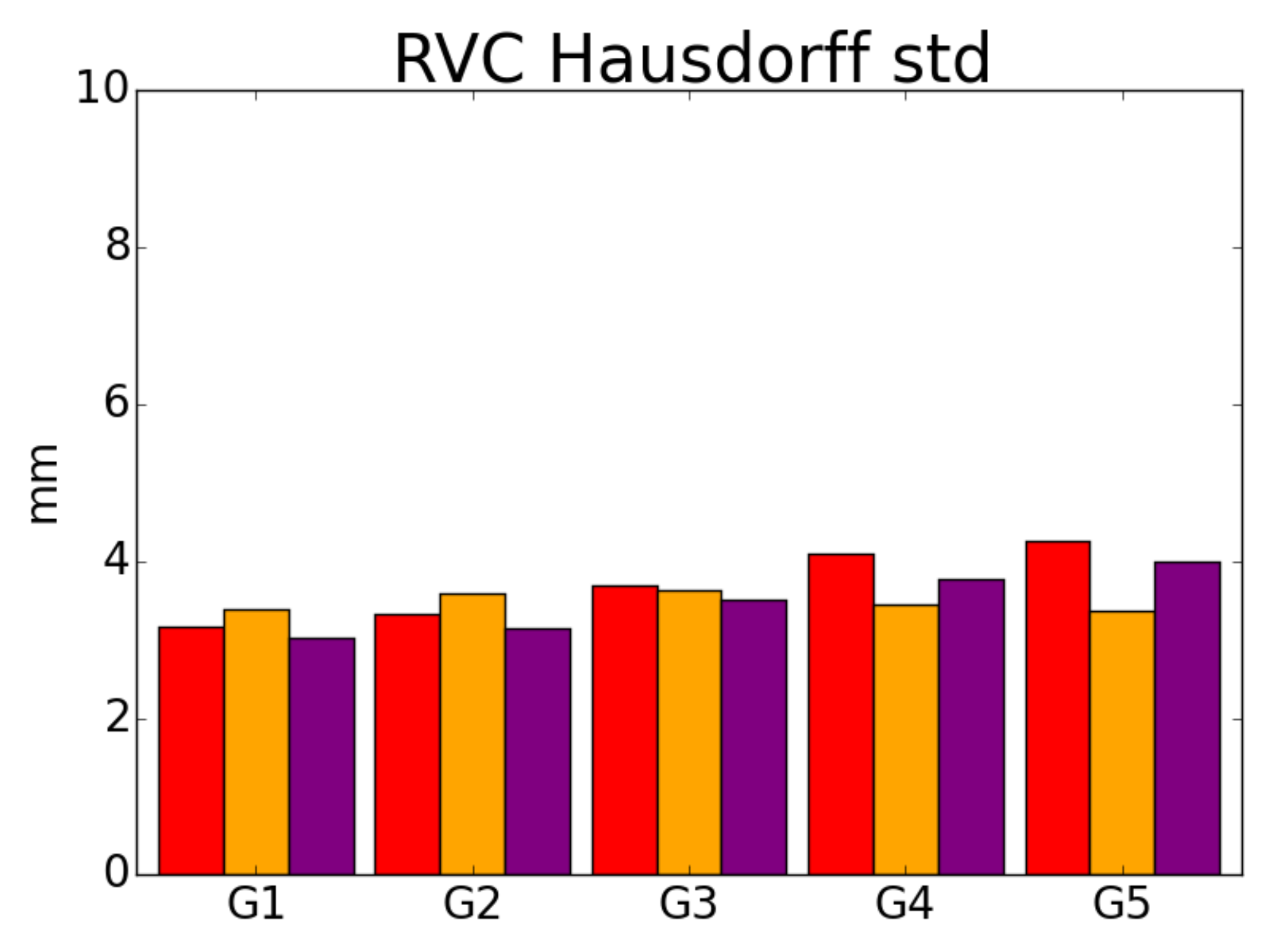}\\

\caption{Performance measured by Hausdorff distance of the LVRV-net, the LVRV-mid-starting-net and the LVRV-no-propagation-net on UK Biobank. The first row indicates both the mean and the standard deviation values, while the second row depicts the standard deviation values only. The four columns stand for LVM, LVC, LV-epi and RVC respectively.}
\end{figure*}

\begin{figure*}[t]
\centering
\includegraphics[width=5.8cm, height=8cm]{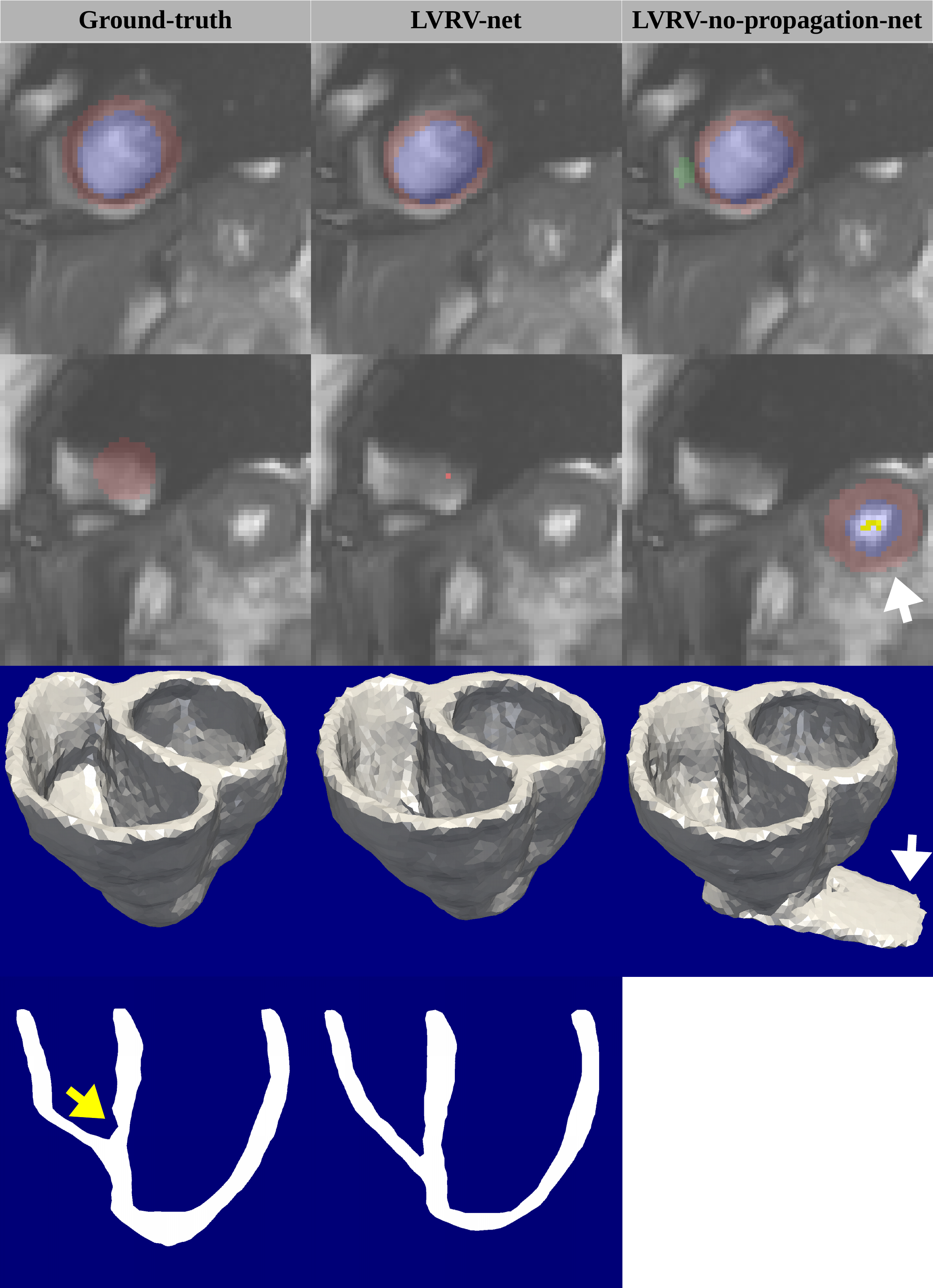}
\hfill
\includegraphics[width=5.8cm, height=8cm]{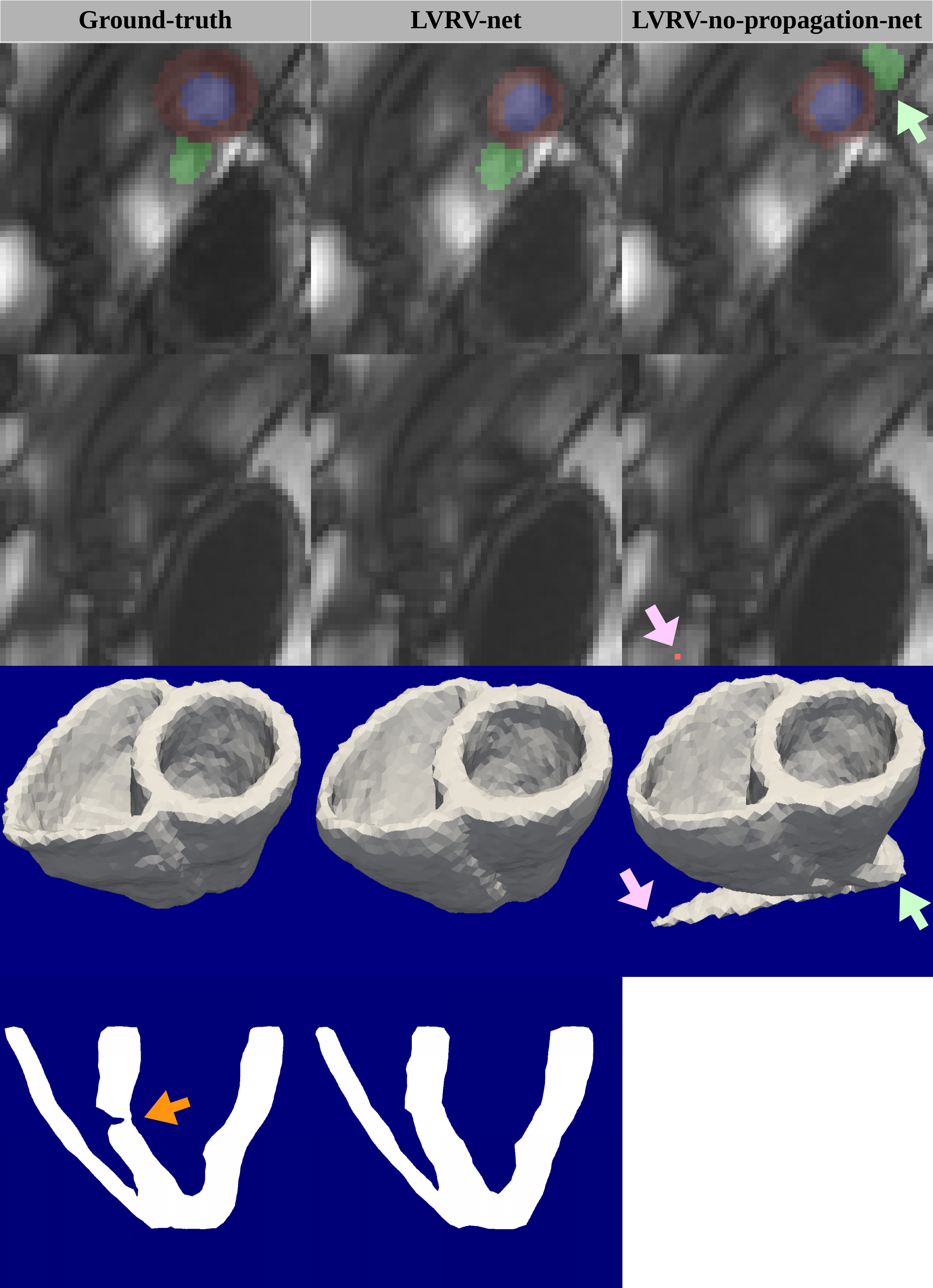}
\hfill
\includegraphics[width=5.8cm, height=8cm]{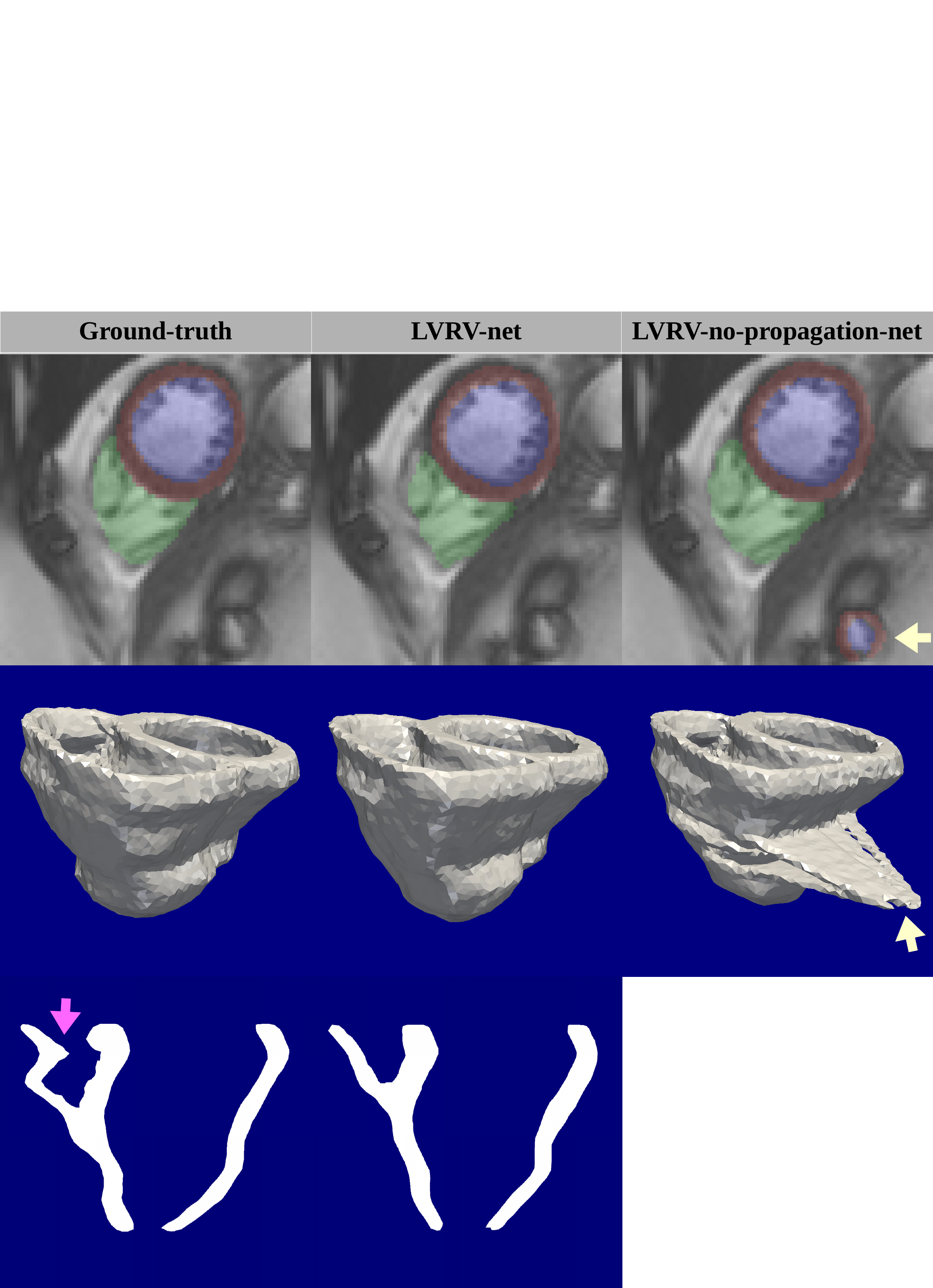}
\caption{Three examples of the segmentation on difficult slices (zoomed-in versions of ROIs for better visualization) and the reconstructed meshes with the ground-truth, the prediction of LVRV-net and that of the LVRV-no-propagation-net. The last row of each example shows a slice of the long-axis view of the meshes reconstructed with the ground-truth and the LVRV-net prediction (irregularities of the ground-truth reconstruction meshes are indicated by the arrows). The large-spread abnormal structures on the meshes in the third column are due to the interpolation of the wrong segmentation (indicated by the arrows).   \quad (Left) The first two rows are the segmentation on the last two slices of the stack. The apex is faint and there is another structure very similar to the heart. The LVRV-net correctly predicts the location of the apex, while the LVRV-no-propagation-net prediction is completely wrong.   \quad (Middle) The first two rows are the segmentation on the last two slices of the stack. The LVRV-no-propagation-net predicts RVC incorrectly on the slice just above the apex and makes a false positive prediction of LVM on the other slice.  \quad (Right) The LVRV-no-propagation-net makes a false positive prediction of LV on an intermediate slice.}
\end{figure*}

The three trained networks are evaluated on the 756 evaluation cases of UK Biobank.
\subsubsection{ROI Determination by ROI-net}
The trained ROI-net is applied to determine and crop ROIs (prediction on the ED sub-stack $S[0.2N, 0.6N]$, the minimum square to cover the union of the predicted masks in the sub-stack, etc.). For all the cases, the determined ROI is successful in the sense that the heart (defined as the union of the pixels labeled to LVC, LVM or RVC in the ground-truth) is fully located inside the ROI, at both ED and ES. Furthermore, all the ROIs are small: the heart and the ROI are distant from $18 \pm 3$ pixels in average, for image and ROI sizes of $209 \pm 1$ and $91 \pm 8$ pixels respectively.

\subsubsection{LVRV-net and LV-net}

We report the segmentation performance in terms of Dice index and Hausdorff distance in Table I. The mean values are reported along with the standard deviation in parentheses. LV-epi is defined as the union of the LVC and LVM.

We notice that the Dice index of the LVM is significantly lower than that of the other parts. We believe that this is partly due to the variability of the ground-truth in UK Biobank as presented in Fig.3. This kind of variability influences both the learning and the evaluation of our method. The Dice index of LVM is most heavily affected. Indeed, on the one hand, LVM is more difficult to segment than LVC due to its shape. Ambiguity on the ground-truth makes the learning of the LVM segmentation even harder. On the other hand, LVM represents a small volume. The Dice index is hence more sensitive to errors in this structure. In general, not only for LVM, the variability in UK Biobank ground-truth reduces the performance for all structures in terms of Dice index. In contrast, the Hausdorff distance is much less sensitive to this variability, which also explains the better performance of our model.

Notice that the results reported in Table I are based on 3D volumes. To evaluate the performance of LVRV-net across different slices, given a structure (e.g. LVM), we also provide results for 5 evenly distributed levels from the slice $S[base$+$1]$ to the last slice on which the structure is present (Fig.4 Left). Group 1 (G1) is on top of the sub-stack and close to the base. Group 5 (G5) is close to the apex. Then we evaluate the segmentation performance of LVRV-net in terms of heart (defined as the union of LVC, LVM, and RVC) presence rate (Fig.4 Right), and 2D Hausdorff distance for 4 different structures (Fig.5).

The evaluation results of LV-net are reported in Table I. Note that although the results of LVRV-net and LV-net are in the same table, LV-net is applied to the basal slice $S[base]$ (the adapted ground-truth of which has no RVC mask) while LVRV-net is not. The higher ground-truth variability on $S[base]$, what we observe in UK Biobank, may explain the slightly lower performance measures of LV-net.

We notice that in general, the performances of the networks are better on ED stacks than on ES stacks. Since the heart is larger at ED than at ES, maybe it is also easier to be segmented at ED.

\subsubsection{LVRV-net vs. Its Variants: Justification of the Top-Starting-Propagation Procedure}
To justify our designs of propagation and of starting propagation from the top slice in the proposed method, we compare LVRV-net with two variants of it, which are considered as baselines. The first baseline is the LVRV-no-propagation-net. Its structure is obtained by removing the extra propagation branch from LVRV-net. So LVRV-no-propagation-net takes an image as its only input and outputs the predicted segmentation mask. LVRV-no-propagation-net is trained and evaluated in the same way as LVRV-net. The evaluation results are reported in Table I, Fig.4 and Fig.5. Another baseline is the LVRV-mid-starting-net. Its structure is identical to that of LVRV-net. But it is trained and then evaluated to segment the middle slice (determined from slice index) in $S[(base$+$1),N]$ with a null contextual input mask, and to propagate the segmentation results upward to the top and down to the bottom of $S[(base$+$1),N]$ using the prediction of the already segmented adjacent slice as the contextual input mask. The results are reported in Table I, Fig.4 and Fig.5.

In Table I, we can see that in terms of Dice index, LVRV-net and LVRV-mid-starting-net are almost the same while LVRV-no-propagation-net is slightly (0.01 to 0.02) higher. Yet in terms of Hausdorff distance, LVRV-net is clearly the best with low values of both mean and standard deviation. Regarding the PR by groups in Fig.4, we find that LVRV-net and LVRV-no-propagation-net detect the presence of the heart slightly better than LVRV-mid-starting-net in G5. In Fig.5 we can see that the differences on mean values of Hausdorff distance are pretty small (within 1mm) for the three networks; but on standard deviation, especially for the LV structures, LVRV-net largely outperforms its variants, sometimes by several mm. Furthermore, we performed the Mann-Whitney $U$ test to prove that the contribution of the propagation is statistically significant. Under the null hypothesis that the LVRV-net and LVRV-no-propagation-net predictions have the same distribution in terms of 3D Hausdorff distance, with the results on the UK Biobank testing set as samples, we obtain $p$-values of $<$0.001, $<$0.001, 0.001, and 0.042 for the LVC, LVM, LV-epi, and RVC respectively, which are small enough ($\leq 0.05$) to conclude on the significance of the results. LVRV-net is clearly more robust than its variants.

To better understand the role of propagation as well as the robustness achieved by the LVRV-net, we look at the cases for which different methods have extremely contrasting performances, and define that the LVRV-no-propagation-net fails while the LVRV-net succeeds on a stack, if the latter outperforms the former on Hausdorff distance by a large value $S$, for any of the 4 structures (LVM, LVC, LV-epi, and RVC). And vice versa. For illustration, we use $S$=30mm, but similar interpretations can stand for other values of $S$. In Fig.6 we  present three typical examples out of the 73 stacks in the UK Biobank testing set for which the LVRV-no-propagation-net fails while the LVRV-net succeeds. In the first example, on the one hand, the apex is so faint on the apical slice that it is barely possible to determine its size precisely. The ground-truth apex seems to be somewhat too large while the LVRV-net prediction looks a little bit too small (in a way learned from the training set with ground-truth variability). But the LVRV-net prediction still well determines the location of the apex using the contextual information. On the other hand, there is a structure on the slice of appearance very similar to the heart. The LVRV-no-propagation-net is confused by it and hence makes a completely wrong prediction. If we reconstruct the anatomical mesh of the heart based on the segmentation (to overcome the problem of large slice thickness, we apply interpolation to generate the segmentation on the intermediate slices between two adjacent slices), the mesh reconstructed from LVRV-no-propagation-net is clearly wrong on the apex. Similarly, the LVRV-no-propagation-net misses the right structure and/or makes a false positive prediction. In contrary, LVRV-net fails while LVRV-no-propagation-net succeeds on only 10 stacks. For 7 of them, LVRV-net predicts a tiny false positive component on a slice either below the apex or around the base. These failures may be simply fixed via the removal of all but the largest connected components. For the other 3 stacks, the errors are caused by image quality problems including large artifact on image and serious misalignment between adjacent slices.

The authors of \cite{Bai:2017} propose a method achieving human-level MRI analysis on UK Biobank. They aim at segmenting as accurately as possible each slice, in contrast with our method, which focuses on the consistency of segmentation across slices. Though the results of their method and that of ours are not directly comparable due to the differences on metrics (e.g. 2D Hausdorff Distance vs 3D Hausdorff Distance), training/testing datasets, preprocessing methods, etc., \cite{Bai:2017} inspired us to compare the performance of our method with that of human experts in terms of 3D consistency, which is the main focus of our method. In Fig.6, for each example, we present a slice of the long-axis view (the last row) for both meshes reconstructed from the ground-truth and the LVRV-net prediction. As indicated by the arrows, with qualitative comparison we find that among these pairs of meshes the ground-truth reconstruction meshes are less regular and less smooth. It suggests that our method maintains 3D consistency even better than human experts.

\subsection{Generalization Ability to Other Datasets}

All the 3 trained networks are applied to the other 3 datasets without finetuning for two reasons. First, we do so to demonstrate their strong generalization ability. Second, as the 3 networks are designed to be big to learn from the large UK Biobank dataset of thousands of cases, finetuning them on small datasets of tens of cases easily results in overfitting. In fact, we have tried to finetune LVRV-net on ACDC. While a certain level (e.g. 10 epochs) of finetuning is beneficial, overfitting happens very soon afterward (obviously since the 50th epoch).

\subsubsection{Experiments on ACDC}

\begin{table*}[]
\caption{Segmentation Results on the ACDC Dataset, Compared to the Performance from the State-of-the-art Methods}
\centering
\begin{tabular}{c cccc cccc}
\hline
\noalign{\vskip 0.0in}
\multicolumn{1}{|}{} & \multicolumn{4}{|c|}{Dice} & \multicolumn{4}{c|}{Hausdorff (mm)} \\ 
 \hline
\multicolumn{1}{|}{} & \multicolumn{2}{|c|}{LVM} & \multicolumn{2}{c|}{LVC} & \multicolumn{2}{c|}{LVM} & \multicolumn{2}{c|}{LVC} \\
\hline
\multicolumn{1}{|}{} & \multicolumn{1}{|c|}{mean} & \multicolumn{1}{c|}{std} & \multicolumn{1}{c|}{mean} & \multicolumn{1}{c|}{std} & \multicolumn{1}{c|}{mean} & \multicolumn{1}{c|}{std} & \multicolumn{1}{c|}{mean} & \multicolumn{1}{c|}{std} \\
\hline

\noalign{\vskip 0.0in}
\multicolumn{1}{|c}{proposed LV-net} & 0.715 & 0.07 & 0.862 & 0.08 & 9.76 & \textbf{3.31} & 8.74 & \multicolumn{1}{c|}{\textbf{3.76}}\\
\multicolumn{1}{|c}{Isensee et al. \cite{Isensee:2017}} & 0.873 & - & 0.930  & - & \textbf{9.668} & - & 8.416 & \multicolumn{1}{c|}{-}\\
\multicolumn{1}{|c}{Jang et al. \cite{Jang:2017}} & \textbf{0.879} & \textbf{0.04} & \textbf{0.938} & \textbf{0.05} & 9.76 & 6.02 & \textbf{7.27} & \multicolumn{1}{c|}{4.83}\\
\multicolumn{1}{|c}{Wolterink et al. \cite{Wolterink:2017}} & 0.87 & \textbf{0.04} & 0.93 & \textbf{0.05} & 11.31 & 5.62 & 8.68 & \multicolumn{1}{c|}{4.51}\\
\hline
\end{tabular}
\end{table*}

\begin{table*}[]
\caption{Segmentation Results by Pathological Group on the ACDC Dataset}
\centering
\begin{tabular}{c cccc cccc}
\hline
\noalign{\vskip 0.0in}
\multicolumn{1}{|}{} & \multicolumn{4}{|c|}{Dice} & \multicolumn{4}{c|}{Hausdorff (mm)} \\ 
 \hline
\multicolumn{1}{|}{} & \multicolumn{2}{|c|}{LVM} & \multicolumn{2}{c|}{LVC} & \multicolumn{2}{c|}{LVM} & \multicolumn{2}{c|}{LVC} \\
\hline
\multicolumn{1}{|}{} & \multicolumn{1}{|c|}{mean} & \multicolumn{1}{c|}{std} & \multicolumn{1}{c|}{mean} & \multicolumn{1}{c|}{std} & \multicolumn{1}{c|}{mean} & \multicolumn{1}{c|}{std} & \multicolumn{1}{c|}{mean} & \multicolumn{1}{c|}{std} \\
\hline

\noalign{\vskip 0.0in}
\multicolumn{1}{|c}{Dilated cardiomyopathy} & 0.705 & \textbf{0.04} & \textbf{0.916} & \textbf{0.02} & \textbf{8.50} & \textbf{2.31} & \textbf{7.19} & \multicolumn{1}{c|}{\textbf{1.81}}\\
\multicolumn{1}{|c}{Hypertrophic cardiomyopathy} & \textbf{0.773} & 0.05 & 0.792  & 0.12 & 12.02 & 3.74 & 11.41 & \multicolumn{1}{c|}{5.48}\\
\multicolumn{1}{|c}{Myocardial infarction} & 0.708 & 0.06 & 0.890 & 0.03 & 9.83 & 3.51 & 8.35 & \multicolumn{1}{c|}{2.40}\\
\multicolumn{1}{|c}{Abnormal right ventricle} & 0.666 & 0.07 & 0.850 & 0.06 & 9.54 & 2.83 & 8.08 & \multicolumn{1}{c|}{2.77}\\
\multicolumn{1}{|c}{Normal} & 0.721 & 0.06 & 0.863 & 0.05 & 8.93 & 2.76 & 8.67 & \multicolumn{1}{c|}{3.72}\\
\hline
\end{tabular}
\end{table*}

The trained ROI-net is applied to the ED sub-stacks $S[0.1N, 0.5N]$ of the 100 ACDC cases. Again as we found in the experiments on UK Biobank, the ROI determination is successful on 100\% of the cases, as all the ROIs contain the heart completely on the one hand, and are very reasonably small on the other hand.

As we pointed out in the ``Data" section, the RVC is segmented in ACDC with conventions quite different from that of UK Biobank. So we only try to segment the LV with the trained LV-net. Some slices to be segmented in ACDC are located well above the base. They are quite different from all the slices used to train LV-net so LV-net can predict some false positives. To deal with this challenge, for the application on ACDC only, we add three more points to the LV-net postprocessing procedure:\\
\textbullet \ The first condition for a successful predicted mask becomes ``both LVC and LVM are present" (instead of ``LVM is present").\\
\textbullet \ If the predicted mask is successful, only the largest components of LVC and LVM are respectively reserved as predicted masks.\\
\textbullet \ If the predicted LVC mask has any neighboring background pixels, we reset the prediction of those pixels to LVC (indicated by the arrows in Fig.7(a)). We do so to follow the ACDC convention that LVC is almost always enclosed by LVC. 

Among the methods in the ACDC challenge, \cite{Isensee:2017} (ranked 1st), \cite{Jang:2017} (ranked 4th) and \cite{Wolterink:2017} (ranked 5th) report their performances on the 100 training cases. The performances on these cases of LV-net and these methods are presented in Table II. Due to the variability of the UK Biobank training set ground-truth, as well as the difference between UK Biobank and ACDC images, LV-net is not as good as the state-of-the-art methods on Dice index. But it is rather comparable to them in terms of the mean of Hausdorff distance, and even better in terms of the standard deviation. This confirms the robustness of our method. In Fig.7(a) we also show some examples of LV-net prediction along with the ACDC ground-truth and the UK Biobank ground-truth on similar slices. It is clear that LV-net learns the segmentation ``pattern" of the ground-truth from UK Biobank, which is different from that of ACDC.

We also find that the difference between the performances of our method on the 5 pathological groups remains limited as presented in Table III. The pathological group seems to have less influence than the image quality of individual stack on the segmentation performance. Being trained with cases from the general population, our method generalizes well to the cases with pathology.

\vspace{2mm}

\subsubsection{Experiments on Sunnybrook}
The slices to be segmented of the 30 cases in Sunnybrook are well located on or below the base of the heart. We segment them with the trained LV-net. In a way similar to the practice in \cite{Tran:2016}, $160\times160$ central zones are cropped out as ROIs, which are then used as inputs to LV-net. Comparison of the performance of LV-net and up-to-date state-of-the-art research is presented in Table IV. LV-net is somewhat less accurate on Dice index and on average perpendicular distance (APD). But its robustness makes it comparable or even better than the state-of-the-art on the percentage of good contours (PGC). Examples of predicted masks and ground-truth are shown in Fig.7(b). 

\begin{table*}
\caption{Segmentation Results on the Sunnybrook Dataset, Compared to the Performance from the State-of-the-art Methods}
\centering
\begin{tabular}{c cccccc cccccc}
\hline
\noalign{\vskip 0.0in}
\multicolumn{1}{|}{} & \multicolumn{4}{|c|}{Dice} & \multicolumn{4}{c|}{APD (mm)} & \multicolumn{4}{c|}{PGC (\%)} \\ 
 \hline
\multicolumn{1}{|}{} & \multicolumn{2}{|c|}{LVC} &\multicolumn{2}{c|}{LV-epi} & \multicolumn{2}{c|}{LVC} & \multicolumn{2}{c|}{LV-epi} & \multicolumn{2}{c|}{LVC} & \multicolumn{2}{c|}{LV-epi} \\
\hline
\multicolumn{1}{|}{} & \multicolumn{1}{|c|}{mean} & \multicolumn{1}{c|}{std} & \multicolumn{1}{c|}{mean} & \multicolumn{1}{c|}{std} & \multicolumn{1}{c|}{mean} & \multicolumn{1}{c|}{std} & \multicolumn{1}{c|}{mean} & \multicolumn{1}{c|}{std} & \multicolumn{1}{c|}{mean} & \multicolumn{1}{c|}{std} & \multicolumn{1}{c|}{mean} & \multicolumn{1}{c|}{std} \\
\hline

\noalign{\vskip 0.0in}
\multicolumn{1}{|c}{proposed LV-net} & 0.88 & 0.07 & 0.94 & 0.03 & 2.11 & 0.49 & 1.95 & 0.42 & 97.08 & 6.04 & \textbf{99.21} & \multicolumn{1}{c|}{2.95} \\
\multicolumn{1}{|c}{Tran \cite{Tran:2016}} & 0.92 & 0.03 & \textbf{0.96} & \textbf{0.01} & \textbf{1.73} & \textbf{0.35} & \textbf{1.65} & \textbf{0.31} & \textbf{98.48} & \textbf{4.06} & 99.17 &\multicolumn{1}{c|}{\textbf{2.20}}\\
\multicolumn{1}{|c}{Winther et al. \cite{Winther:2017}} & \textbf{0.94} & 0.03 & 0.95 & 0.03 & - & - & - & - & - & - & - & \multicolumn{1}{c|}{-}\\
\multicolumn{1}{|c}{Avendi et al. \cite{Avendi:2016}} & \textbf{0.94} & \textbf{0.02} & - & - & 1.81 & 0.44 & - & - & 96.69 & 5.7 & - & \multicolumn{1}{c|}{-}\\
\multicolumn{1}{|c}{Queiros et al. \cite{Queiros:2014}} & 0.90 & 0.05 & 0.94 & 0.02 & 1.76 & 0.45 & 1.80 & 0.41 & 92.70 & 9.5 & 95.40 & \multicolumn{1}{c|}{9.6}\\
\multicolumn{1}{|c}{Poudel et al. \cite{Poudel:2016}} & 0.90 & 0.04 & - & - & 2.05 & \textbf{0.29} & - & - & 95.34 & 7.2 & - & \multicolumn{1}{c|}{-}\\
\hline
\end{tabular}
\end{table*}

\vspace{2mm}

\subsubsection{Experiments on RVSC}
The slices to be segmented for the 16 cases in RVSC are all located below the base and above the apex. Similar to \cite{Tran:2016}, $216\times216$ central zones are cropped out as ROIs. We then apply the trained LVRV-net on these ROIs and evaluate the predicted RVC masks. Comparison with the up-to-date state-of-the-art research is presented in Table V. In terms of Hausdorff distance, our method not only achieves better mean value but also generates much smaller standard deviation value compared the to state-of-the-art. Examples of predicted masks and ground-truth are presented in Fig.7(c). 

\begin{table}[]
\caption{Segmentation Results on the RVSC Dataset, Compared to the Performance from the State-of-the-art Methods}
\centering
\begin{tabular}{c cc cc}
\hline
\noalign{\vskip 0.0in}
\multicolumn{1}{|}{} & \multicolumn{2}{|c|}{Dice} & \multicolumn{2}{c|}{Hausdorff (mm)} \\ 
\hline
\multicolumn{1}{|}{} & \multicolumn{2}{|c|}{RVC} &\multicolumn{2}{c|}{RVC} \\
\hline
\multicolumn{1}{|}{} & \multicolumn{1}{|c|}{mean} & \multicolumn{1}{c|}{std} & \multicolumn{1}{c|}{mean} & \multicolumn{1}{c|}{std} \\
\hline
\noalign{\vskip 0.0in}
\multicolumn{1}{|c}{proposed LVRV-net} & 0.82  & \textbf{0.07} & \textbf{7.56} & \multicolumn{1}{c|}{\textbf{3.50}}\\
\multicolumn{1}{|c}{Tran \cite{Tran:2016}} & 0.84 & 0.21 & 8.86 & \multicolumn{1}{c|}{11.27} \\
\multicolumn{1}{|c}{Winther et al. \cite{Winther:2017}} & \textbf{0.85} & \textbf{0.07} & - & \multicolumn{1}{c|}{-} \\
\multicolumn{1}{|c}{Avendi et al. \cite{Avendi:2016:2}} & 0.81 & 0.21 & 7.79 & \multicolumn{1}{c|}{5.91} \\
\multicolumn{1}{|c}{Zuluaga et al. \cite{Zuluaga:2013}} & 0.76 & 0.25 & 11.51 & \multicolumn{1}{c|}{10.06} \\
\hline
\end{tabular}
\end{table}

\begin{figure*}[]
\centering
\subfigure[ACDC dataset (the arrows indicate the pixel labels reset to LVM)]{
\includegraphics[width=5.7cm, height=5.5cm]{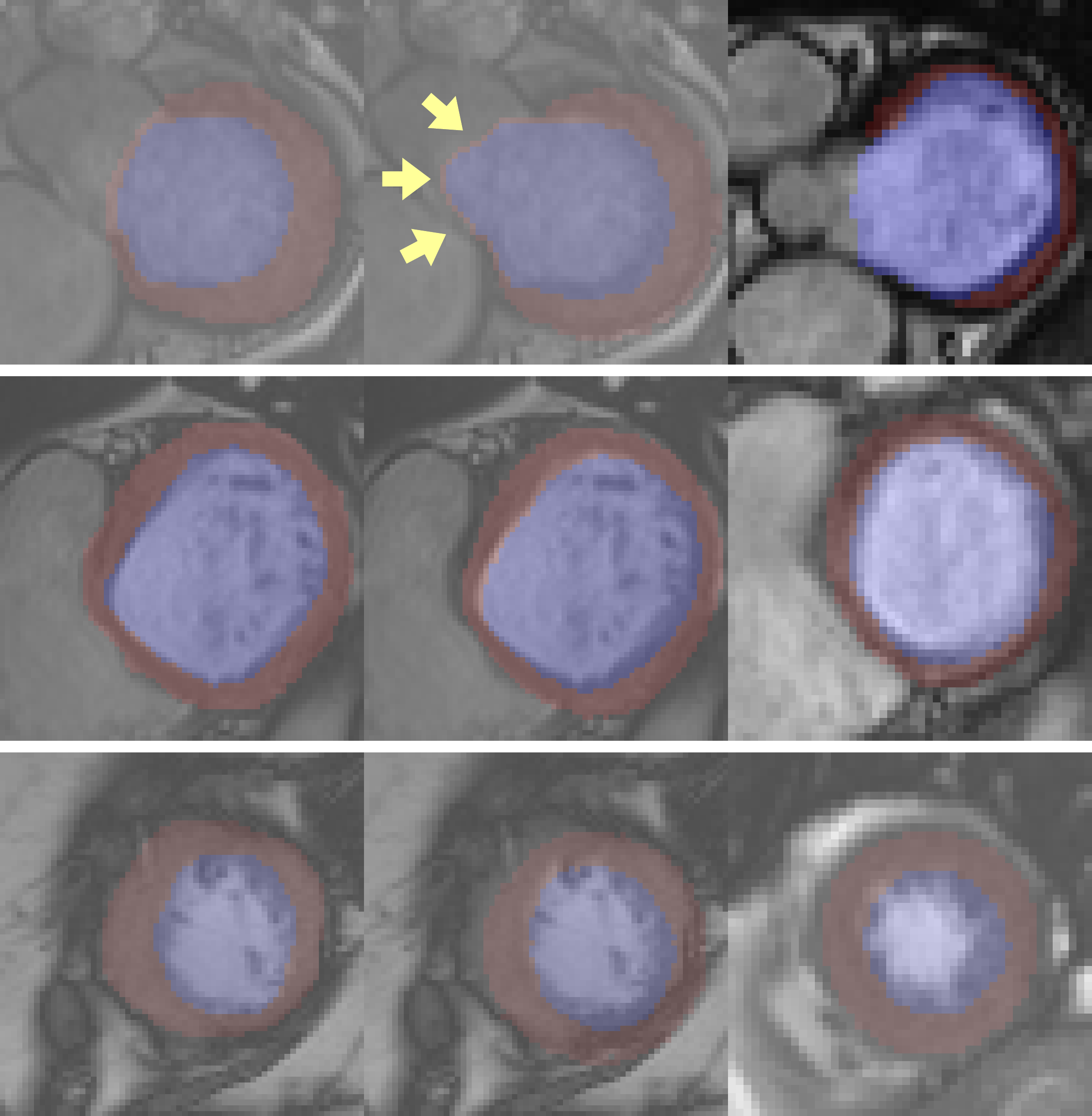}}\quad\quad
\subfigure[Sunnybrook dataset]{
\includegraphics[width=3.8cm, height=5.5cm]{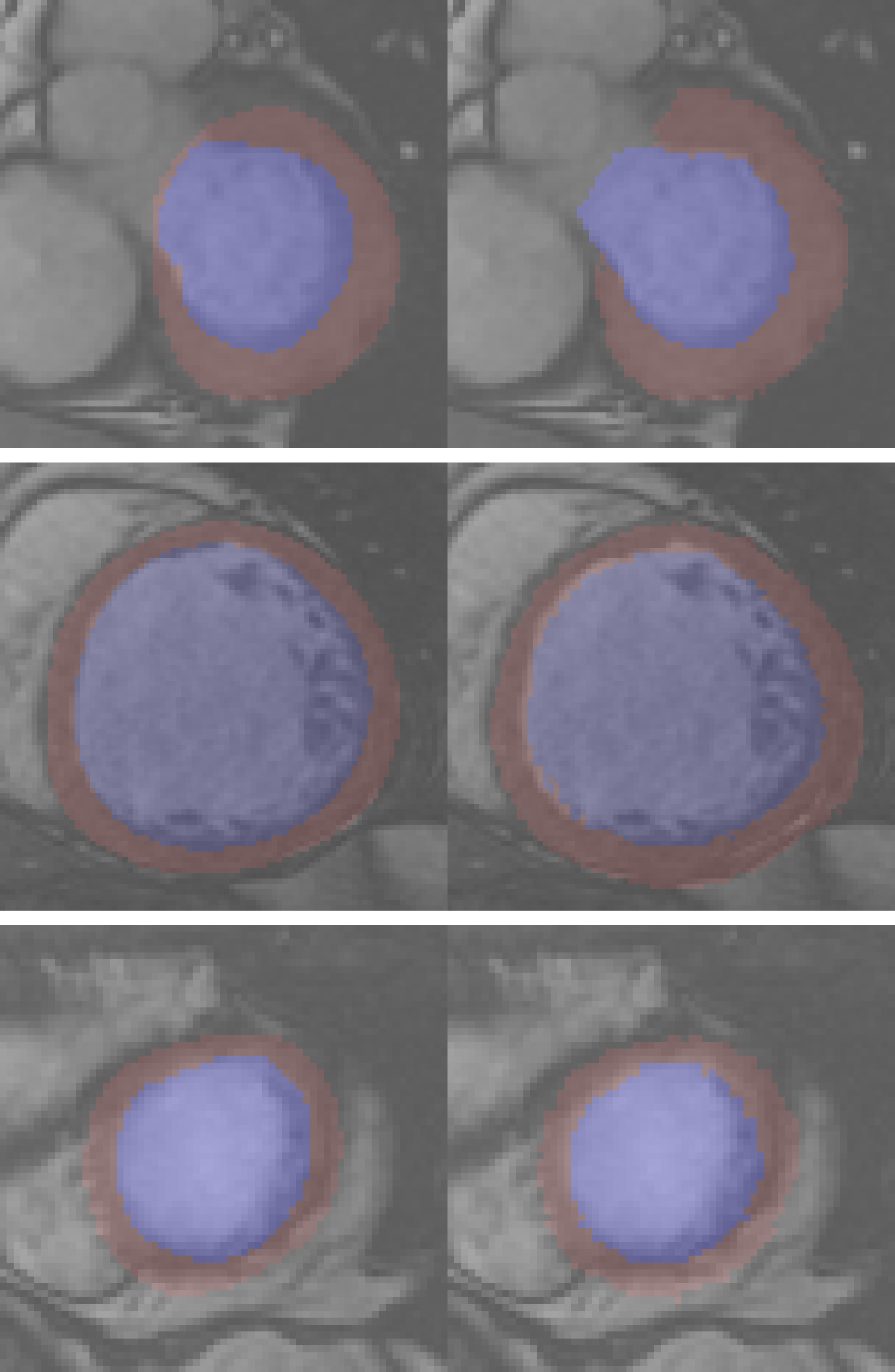}}\quad\quad
\subfigure[RVSC dataset]{
\includegraphics[width=3.8cm, height=5.5cm]{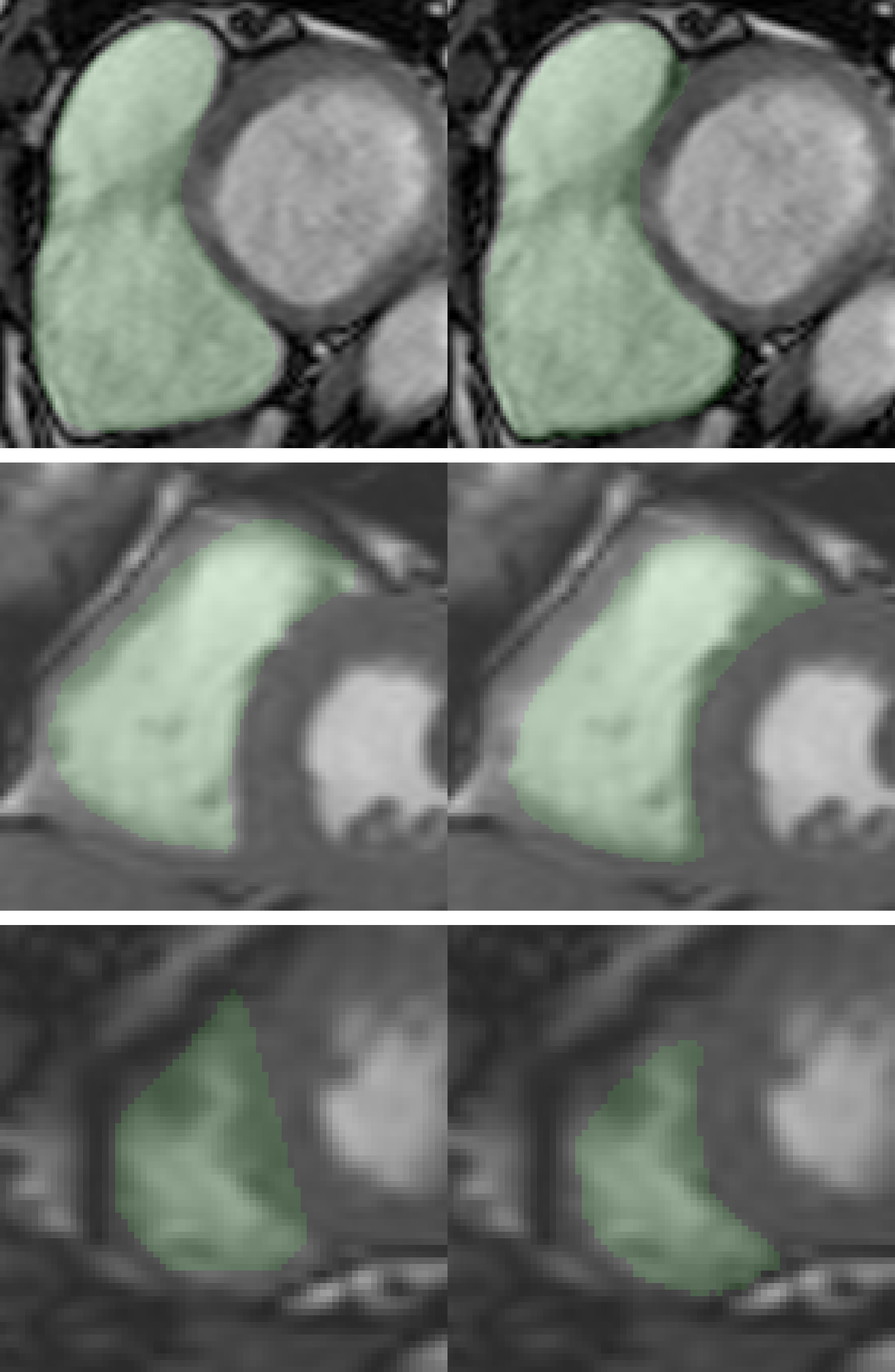}}
\caption{Examples of ground-truth (first column) vs prediction (second column) on 3 datasets. In (a) we also add similar slices with the ground-truth in UK Biobank (third column). The 3 rows correspond to slices roughly on the top (around the base), in the middle and at the bottom (around the apex) of image stacks. LVC, LVM and RVC are marked as purple, brown and green respectively. Note that these images are zoomed-in versions of ROIs for the sake of better visualization. They are not the ROIs which LVRV-net and LV-net take as inputs.}
\end{figure*}

\section{Conclusion and Discussion}

We propose a method of segmentation with spatial propagation that is based on originally designed neural networks. By taking the contextual input into account, the spatial consistency of segmentation is enforced. Also, we conduct thorough and unprecedented testing to evaluate the generalization ability of our model and achieve performance better than or comparable to the state-of-the-art. Furthermore, an exceptionally large dataset (UK Biobank) collected from the general population is used for training and evaluation.

Given the experiments in this paper, we notice that our method is very robust in terms of distance measures (e.g. Hausdorff distance) but less precise than the state-of-the-art in terms of Dice index. The variability of ground-truth in the UK Biobank training set is one important reason for that. For instance, the high ground-truth variability on the basal slice, which is included in the testing
sub-stacks for LV-net but not for LVRV-net, explains the slightly lower performance measures of LV-net in Table I. Yet this kind of variability commonly exists in large datasets so we have to decide to accept and cope with it. Furthermore, inconsistency problems may occur in segmentation (as illustrated and discussed), to which the Dice index might not be sensitive. We believe that on this problem more attention should be paid to the Hausdorff distance, according to which our proposed method performs better. For instance, in the third example shown in Fig.6, a small spot of false positive of LVC segmentation is predicted by LVRV-no-propagation-net. This is a very typical case of inconsistency: the false positive part is quite small compared to the ground-truth LVC, and therefore only causes a slight reduction of the Dice index. But it certainly brings about an explosion of the Hausdorff distance.

We did not directly measure the human performance in terms of 3D metrics on UK Biobank to compare with our method. However, the authors of \cite{Bai:2017} did conduct experiments on UK Biobank to measure human performance in terms of 2D metrics. Taking the inter-observer variability of 3 human experts into account, the reported human expert levels are about 0.93(LVC), 0.88(LVM), and 0.88(RVC) in terms of 2D Dice index, and about 3.1mm(LVC), 3.8mm(LVM) and 7.4mm(RVC) in terms of 2D Hausdorff distance. Though these results are not directly comparable to ours, they may still give a rough idea of human performance. We roughly estimate that our method, while mainly focusing on consistency, has a performance still a little bit lower than that of human experts in terms of accuracy.

Most of the existing segmentation methods do not explicitly take spatial consistency into account. In particular, they do not accurately segment the ``difficult" slices around the apex. Our method, segmenting in a spatially consistent manner, is particularly more robust than them on these slices. The importance of correctly segmenting these slices is often underestimated. In many cutting-edge research projects (e.g. cardiac motion simulation and image synthesis), as a primary step, 3D meshes need to be built based on segmentation. Without spatial consistency and success on the apical slices of the segmentation, the generated meshes would be problematic.

Finally, we wonder whether our method, with better performance on distance measures than many state-of-the-art methods, would be a great tool for cardiac motion analysis. Intuitively, the smaller the Hausdorff distance between the predicted and the ground-truth contours at each instant is, the more precisely the trajectory of the corresponding structure (e.g. LVC, LVM, RVC) can be tracked across time, and hence the better the motion can be characterized. We expect to carry out research on this in the future.

\section*{Acknowledgments}

{\footnotesize The authors acknowledge the partial support from the European Research Council (MedYMA ERC-AdG-2011-291080). We particularly thank Steffen Petersen for giving us access to the UK Biobank study (ANID 2964, PI: S. Petersen, funded by British Heart Foundation (PG/14/89/31194)), and Daniel Rueckert and Wenjia Bai for providing the tools to convert the UK Biobank ground-truth data into a common format. We also thank all the contributors who created the ground-truth segmentation data attached to the UK Biobank cardiac images (cf. reference \cite{Bai:2017}).}

\bibliographystyle{IEEEtran}
\bibliography{ref}

\begin{thebibliography}{10}
\providecommand{\url}[1]{#1}
\csname url@samestyle\endcsname
\providecommand{\newblock}{\relax}
\providecommand{\bibinfo}[2]{#2}
\providecommand{\BIBentrySTDinterwordspacing}{\spaceskip=0pt\relax}
\providecommand{\BIBentryALTinterwordstretchfactor}{4}
\providecommand{\BIBentryALTinterwordspacing}{\spaceskip=\fontdimen2\font plus
\BIBentryALTinterwordstretchfactor\fontdimen3\font minus
  \fontdimen4\font\relax}
\providecommand{\BIBforeignlanguage}[2]{{%
\expandafter\ifx\csname l@#1\endcsname\relax
\typeout{** WARNING: IEEEtran.bst: No hyphenation pattern has been}%
\typeout{** loaded for the language `#1'. Using the pattern for}%
\typeout{** the default language instead.}%
\else
\language=\csname l@#1\endcsname
\fi
#2}}
\providecommand{\BIBdecl}{\relax}
\BIBdecl

\bibitem{Petersen:2016}
S.~Petersen, P.~Matthews, J.~Francis, M.~Robson, F.~Zemrak, R.~Boubertakh,
  A.~Young, S.~Hudson, P.~Weale, S.~Garratt, R.~Collins, S.~Piechnik, and
  S.~Neubauer, ``{UK} {B}iobank's cardiovascular magnetic resonance protocol,''
  \emph{J Cardiovasc Magn Reson}, vol. 18:8, pp.~8+, 2016.

\bibitem{Winther:2017}
H.~Winther, C.~Hundt, B.~Schmidt, C.~Czerner, J.~Bauersachs, F.~Wacker, and
  J.~Vogel-Claussen, ``$\nu$-net: Deep learning for generalized biventricular
  cardiac mass and function parameters,'' \emph{arXiv preprint
  arXiv:1706.04397}, 2017.

\bibitem{Tran:2016}
P.~Tran, ``A fully convolutional neural network for cardiac segmentation in
  short-axis {MRI},'' \emph{arXiv preprint arXiv:1604.00494}, 2016.

\bibitem{Baumgartner:2017}
C.~Baumgartner, L.~Koch, M.~Pollefeys, and E.~Konukoglu, ``An exploration of
  2{D} and 3{D} deep learning techniques for cardiac {MR} image segmentation,''
  in \emph{Proc. Statistical Atlases and Computational Models of the Heart
  (STACOM), ACDC challenge, MICCAI'17 Workshop}, 2017.

\bibitem{Isensee:2017}
F.~Isensee, P.~Jaeger, P.~Full, I.~Wolf, S.~Engelhardt, and K.~Maier-Hein,
  ``Automatic cardiac disease assessment on cine-{MRI} via time-series
  segmentation and domain specific features,'' in \emph{Proc. Statistical
  Atlases and Computational Models of the Heart (STACOM), ACDC challenge,
  MICCAI'17 Workshop}, 2017.

\bibitem{Poudel:2016}
R.~Poudel, P.~Lamata, and G.~Montana, ``Recurrent fully convolutional neural
  networks for multi-slice {MRI} cardiac segmentation,'' \emph{arXiv preprint
  arXiv:1608.03974}, 2016.

\bibitem{Ronneberger:2015}
O.~Ronneberger, P.~Fischer, and T.~Brox, ``U-net: Convolutional networks for
  biomedical image segmentation,'' \emph{MICCAI}, vol. 9351, pp. 234--241,
  2015.

\bibitem{Suinesiaputra:2015}
A.~Suinesiaputra, D.~Bluemke, B.~Cowan, M.~Friedrich, C.~Kramer, R.~Kwong,
  S.~Plein, J.~Schulz-Menger, J.~Westenberg, A.~Young, and E.~Nagel,
  ``Quantification of {LV} function and mass by cardiovascular magnetic
  resonance: multi-center variability and consensus contours,'' \emph{Journal
  of Cardiovascular Magnetic Resonance}, vol. 17(1), 2015.

\bibitem{Oktay:2018}
O.~Oktay, E.~Ferrante, K.~Kamnitsas, M.~Heinrich, W.~Bai, J.~Caballero,
  S.~Cook, A.~de~Marvao, T.~Dawes, D.~O'Regan, B.~Kainz, B.~Glocker, and
  D.~Rueckert, ``Anatomically constrained neural networks ({ACNN}s):
  application to cardiac image enhancement and segmentation,'' \emph{IEEE Trans
  Med Imaging}, vol.~37, pp. 384--395, 2018.

\bibitem{Radau:2009}
P.~Radau, Y.~Lu, K.~Connelly, G.~Paul, A.~Dick, and G.~Wright, ``Evaluation
  framework for algorithms segmenting short axis cardiac {MRI},'' \emph{The
  MIDAS Journal - Cardiac MR Left Ventricle Segmentation Challenge
  \url{http://hdl.handle.net/10380/3070}}, 2009.

\bibitem{Petitjean:2015}
C.~Petitjean, M.~Zuluaga, W.~Bai, J.~Dacher, D.~Grosgeorge, J.~Caudron,
  S.~Ruan, I.~Ayed, M.~Cardoso, H.~Chen, D.~JimenezCarretero,
  M.~Ledesma-Carbayo, C.~Davatzikos, J.~Doshi, G.~Erus, O.~Maier, C.~Nambakhsh,
  Y.~Ou, S.~Ourselin, C.~Peng, N.~Peters, T.~Peters, M.~Rajchl, D.~Rueckert,
  A.~Santos, W.~Shi, C.~Wang, H.~Wang, and J.~Yuan, ``Right ventricle
  segmentation from cardiac {MRI}: A collation study,'' \emph{Medical Image
  Analysis}, vol. 19(1), pp. 187--202, 2015.

\bibitem{Schulz-Menger:2013}
J.~Schulz-Menger, D.~Bluemke, J.~Bremerich, S.~Flamm, M.~Fogel, M.~Friedrich,
  R.~Kim, F.~von Knobelsdorff-Brenkenhoff, C.~Kramer, D.~Pennell, S.~Plein, and
  E.~Nagel, ``Standardized image interpretation and post processing in
  cardiovascular magnetic resonance: society for cardiovascular magnetic
  resonance ({SCMR}) board of trustees task force on standardized post
  processing,'' \emph{J Cardiovasc Magn Reson}, vol. 15(35), pp. 1167--1186,
  2013.

\bibitem{Maas:2013}
A.~Maas, A.~Hannun, and A.~Ng, ``Rectifier nonlinearities improve neural
  network acoustic models,'' \emph{Proc. ICML}, vol.~30, 2013.

\bibitem{Kayalibay:2017}
B.~Kayalibay, G.~Jensen, and P.~van~der Smagt, ``{CNN}-based segmentation of
  medical imaging data,'' \emph{arXiv preprint arXiv:1701.03056}, 2017.

\bibitem{Wolterink:2017}
J.~Wolterink, T.~Leiner, M.~Viergever, and I.~Isgum, ``Automatic segmentation
  and disease classification using cardiac cine {MR} images,'' in \emph{Proc.
  Statistical Atlases and Computational Models of the Heart (STACOM), ACDC
  challenge, MICCAI'17 Workshop}, 2017.

\bibitem{Bai:2017}
W.~Bai, M.~Sinclair, G.~Tarroni, O.~Oktay, M.~Rajchl, G.~Vaillant, A.~Lee,
  N.~Aung, E.~Lukaschuk, M.~Sanghvi, F.~Zemrak, K.~Fung, J.~Paiva,
  V.~Carapella, Y.~Kim, H.~Suzuki, B.~Kainz, P.~Matthews, S.~Petersen,
  S.~Piechnik, S.~Neubauer, B.~Glocker, and D.~Rueckert, ``Human-level {CMR}
  image analysis with deep fully convolutional networks,'' \emph{arXiv preprint
  arXiv:1710.09289}, 2017.

\bibitem{Jang:2017}
Y.~Jang, S.~Ha, S.~Kim, Y.~Hong, and H.~Chang, ``Automatic segmentation of {LV}
  and {RV} in cardiac {MRI},'' in \emph{Proc. Statistical Atlases and
  Computational Models of the Heart (STACOM), ACDC challenge, MICCAI'17
  Workshop}, 2017.

\bibitem{Avendi:2016}
M.~Avendi, A.~Kheradvar, and H.~Jafarkhani, ``A combined deeplearning and
  deformable-model approach to fully automatic segmentation of the left
  ventricle in cardiac {MRI},'' \emph{Med Image Anal}, vol.~30, pp. 108--109,
  2016.

\bibitem{Queiros:2014}
S.~Queiros, D.~Barbosa, B.~Heyde, P.~Morais, J.~Vilaca, D.~Friboulet,
  O.~Bernard, and J.~D'hooge, ``Fast automatic myocardial segmentation in {4D}
  cine {CMR} datasets,'' \emph{Med Image Anal}, vol.~18, pp. 1115--1131, 2014.

\bibitem{Avendi:2016:2}
M.~Avendi, A.~Kheradvar, and H.~Jafarkhani, ``Fully automatic segmentation of
  heart chambers in cardiac {MRI} using deep learning,'' \emph{J Cardiovasc
  Magn Reson}, vol.~18, pp. 351--353, 2016.

\bibitem{Zuluaga:2013}
M.~Zuluaga, M.~Cardoso, M.~Modat, and S.~Ourselin, ``Multi-atlas propagation
  whole heart segmentation from {MRI} and {CTA} using a local normalised
  correlation coefficient criterion,'' \emph{Functional Imaging and Modeling of
  the Heart}, pp. 172--180, 2013.

\end{thebibliography}

\appendix

\section{}

\subsection{Datasets}
\subsubsection{UK Biobank Dataset}

It comprises short-axis cine MRI of 4875 participants from the general population. Details of the magnetic resonance protocol are described in \cite{Petersen:2016}. Each time series is composed of 3D volumes with 10mm slice thickness and in-plane resolution ranging from 1.8mm to 2.3mm. Expert manual segmentation using CVI42 \footnote{\url{https://www.circlecvi.com/}} for LVC, LVM, and RVC is provided as ground-truth at both ED and ES. The quality of ground-truth varies highly across the cases. We exclude about one thousand cases that are provided with incomplete (e.g. missing ground-truth on some slice(s)) or unconvincing ground-truth (e.g. visually significant image/mask mismatch). Then we split the remaining 3834 cases into 2 sets of 3078 cases and 756 cases, for training and evaluation respectively.

\subsubsection{Automated Cardiac Diagnosis Challenge (ACDC) Dataset}

The ACDC dataset comprises short-axis cine MRI of 100 subjects, which are divided into 5 groups of equal size: dilated cardiomyopathy, hypertrophic cardiomyopathy, myocardial infarction, abnormal right ventricle and normal subjects. Each time series is composed of 3D volumes with 5mm to 10mm  slice thickness and in-plane resolution ranging from 0.7mm to 1.9mm. Expert manual segmentation for LVC, LVM, and RVC is provided as ground-truth at both ED and ES phases.

\subsubsection{Sunnybrook Dataset}

The validation and the online sub-datasets of the Sunnybrook dataset, made available for the MICCAI 2009 challenge on automated left ventricle (LV) segmentation, contains short-axis cine MRI from 30 subjects with different cardiac conditions: healthy (6 cases), hypertrophy (8 cases), heart failure with infarction (8 cases), and heart failure without infarction (8 cases). Each time series is composed of 6 to 12 2D cine stacks with a slice thickness of 8mm and in-plane resolution ranging from 1.3mm to 1.4mm. Expert-delineated ground-truth contours of the endocardium, or LVC, are provided at both ED and ES phases. Those of epicardium, or LVM, are provided only at ED phase.

\subsubsection{Right Ventricle Segmentation Challenge (RVSC) Dataset}

The RVSC dataset comprises 16 training 2D short-axis cine MRI stacks consisting of slices located across the ventricle. The in-plane resolution ranges from 0.57mm to 0.97mm. Ground-truth delineation of endocardial borders (LVC contours) and epicardial borders are provided at both ED and ES phases for the training cine stacks.

\subsection{Metrics}
\subsubsection{Dice Index}
The Dice index measures the overlap between two areas (2D Dice index) or two volumes (3D Dice index). It is defined as
\begin{equation}
\mathcal{D}(A,B)= 2\frac{A \cap B}{A + B}
\end{equation}
for two areas or two volumes $A$ and $B$. The Dice index varies from 0 (complete mismatch) to 1 (perfect match).
\subsubsection{Hausdorff Distance}
The Hausdorff distance measures the distance between two areas (2D Hausdorff distance) or two volumes (3D Hausdorff distance). It is defined as
\begin{equation}
\mathcal{H}(A,B)= \max \Big( \max_{p \in A}\big(\min_{q \in B}d(p,q) \hspace{0.5ex} \big), \hspace{0.5ex} \max_{q \in B}\big(\min_{p \in A}d(p,q) \hspace{0.5ex} \big) \Big)
\end{equation}
where $d$ denotes Euclidean distance. A smaller Hausdorff distance implies a better match.
\subsubsection{Average Perpendicular Distance}
The average perpendicular distance (APD) \cite{Radau:2009} measures the distance in mm from one contour to another, averaged over all contour points.
\subsubsection{Percentage of Good Contours}
Given a set of ground-truth contours and the corresponding predicted contours, the percentage of good contours (PGC) defined in \cite{Radau:2009} is the fraction of the predicted contours which have APD less than 5mm away from the ground-truth contours.
\subsubsection{Presence Rate}
Segmentation methods may miss a structure totally on some difficult slices. Given the segmentation predictions on a sub-stack, the presence rate (PR) of a structure is defined as the ratio between the number of predicted masks with the structure and the number of slices in the sub-stack.

On the UK Biobank and ACDC datasets, we use the 3D Dice index and 3D Hausdorff distance as metrics, similar to what has been done for the ACDC STACOM MICCAI 2017 challenge. For Sunnybrook, we use the 2D Dice index, APD, and PGC as in the MICCAI 2009 challenge on automated LV segmentation. For RVSC, we use 2D Dice index and 2D Hausdorff distance as done for the MICCAI 2012 challenge on automated RV segmentation.

\end{document}